\documentclass{article} % For LaTeX2e
\usepackage[utf8]{inputenc} % allow utf-8 input
\usepackage[T1]{fontenc}    % use 8-bit T1 fonts
\usepackage{geometry} % to change the page dimensions
\usepackage{natbib}
\usepackage{fancyhdr} 
\usepackage{amsthm}
\usepackage{graphicx}

\usepackage{amsmath,amsfonts,bm}

\def\eqref#1{equation~\ref{#1}}
\def\1{\bm{1}}

\def\vh{{\bm{h}}}

\def\vp{{\bm{p}}}

\def\vt{{\bm{t}}}

\def\vv{{\bm{v}}}

\def\vy{{\bm{y}}}
\def\vz{{\bm{z}}}

\def\mU{{\bm{U}}}

\def\mW{{\bm{W}}}

\DeclareMathAlphabet{\mathsfit}{\encodingdefault}{\sfdefault}{m}{sl}
\SetMathAlphabet{\mathsfit}{bold}{\encodingdefault}{\sfdefault}{bx}{n}

\newcommand{\R}{\mathbb{R}}

\usepackage[colorlinks=true]{hyperref}
\usepackage{url}
\usepackage{cleveref}
\usepackage{xcolor}
\usepackage[textwidth=1in,textsize=tiny]{todonotes}
\usepackage{multicol,multirow}
\usepackage{wrapfig}
\usepackage{mathtools}
\usepackage{amsmath,amsfonts,amssymb,amsthm,commath}

\usepackage{graphicx}
\graphicspath{ {./figures/} }
\usepackage{caption}
\usepackage{multirow}
\usepackage{booktabs}
\usepackage{subcaption}
\usepackage{listings}
\usepackage{diagbox}
\usepackage[inline]{enumitem}

\def\ddefloop#1{\ifx\ddefloop#1\else\ddef{#1}\expandafter\ddefloop\fi}
\def\ddef#1{\expandafter\def\csname bb#1\endcsname{\ensuremath{\mathbb{#1}}}}
\ddefloop ABCDEFGHIJKLMNOPQRSTUVWXYZ\ddefloop
\def\ddef#1{\expandafter\def\csname c#1\endcsname{\ensuremath{\mathcal{#1}}}}
\ddefloop ABCDEFGHIJKLMNOPQRSTUVWXYZ\ddefloop

\newcommand*\subtxt[1]{_{\textnormal{#1}}}
\DeclareRobustCommand\_{\ifmmode\expandafter\subtxt\else\textunderscore\fi}

\DeclarePairedDelimiter\autobracket{(}{)}
\newcommand{\p}[1]{\autobracket*{#1}}

\newcommand{\ip}[2]{\left\langle #1, #2 \right \rangle}
\theoremstyle{definition}
\title{\textsc{Text Descriptions are Compressive and Invariant Representations for Visual Learning}}

\author{Zhili Feng \\
Carnegie Mellon University\\
\texttt{zhilif@andrew.cmu.edu} \\
\and
Anna Bair\\
Carnegie Mellon University\\
\texttt{abair@cmu.edu}\\
\and
J. Zico Kolter \\
Carnegie Mellon University \\
Bosch Center for AI\\
\texttt{zkolter@cs.cmu.edu}\\
}

\begin{document}

\maketitle

\begin{abstract}

Modern image classification is based upon directly predicting classes via large discriminative networks, which do not directly contain information about the intuitive visual features that may constitute a classification decision. 
Recently, work in vision-language models (VLM) such as CLIP has provided ways to specify natural language descriptions of image classes, but typically focuses on providing single descriptions for each class. In this work, we demonstrate that an alternative approach, in line with humans' understanding of multiple visual features per class, can also provide compelling performance in the robust few-shot learning setting. 
In particular, we introduce a novel method, \textit{SLR-AVD (Sparse Logistic Regression using Augmented Visual Descriptors)}. This method first automatically generates multiple visual descriptions of each class via a large language model (LLM), then uses a VLM to translate these descriptions to a set of visual feature embeddings of each image, and finally uses sparse logistic regression to select a relevant subset of these features to classify each image.
Core to our approach is the fact that, information-theoretically, these descriptive features are more invariant to domain shift than traditional image embeddings, even though the VLM training process is not explicitly designed for invariant representation learning. These invariant descriptive features also compose a better input compression scheme. When combined with finetuning, we show that SLR-AVD is able to outperform existing state-of-the-art finetuning approaches on both in-distribution and out-of-distribution performance.

\end{abstract}
\section{Introduction}
Self-supervised vision-language models (VLMs) like CLIP \citep{radford2021learning} create aligned image and text encoders via contrastive training.  Unlike traditionally-trained classification networks, such alignment enables zero-shot image classification by prompting the text encoder with hand-crafted inputs like ``\texttt{a photo of \{\textsl{}\}}'' then predicting the target via the maximal inner product with the input image embedding.  However, choosing effective prompts for zero-shot learning remains largely an ad-hoc process: \citet{radford2021learning} has added several prompts like ``\texttt{the cartoon \{\}}'' or ``\texttt{art of the \{\}}'' aiming to improve ImageNet-R \citep{hendrycks2021many} performance, which (somewhat surprisingly) improved standard ImageNet accuracy as well.  This has led to works that attempt to automatically extract relevant prompts from language models \citep{pratt2022does}, including work that uses these models to extract \emph{multiple} visual descriptors \citep{menon2022visual} then use the average prediction of these visual descriptions to classify the image.

In the few-shot setting, however, where a small amount of training data is available, a number of techniques can further improve classifier performance beyond zero-shot prompting alone.  For example, it has become commonplace to finetune zero-shot classifiers via linear probing or other approaches \citep{kumar2022fine}, including methods that interpolate between the zero-shot and finetuned classifiers \citep{wortsman2022robust} to achieve better out-of-distribution robustness.  Alternatively, one can also adapt the prompts themselves using this few-shot data, using e.g. techniques from soft prompt tuning \citep{zhou2022learning}, though these learned prompts are not readable, nor are their nearest dictionary projections \citep{khashabi2021prompt}.  Finally, recent work has also looked at ways to combine automatically-extracted prompts using few-shot learning \citep{yang2022language}, though this approach used a very specific learned weighting over such descriptions for interpretability purposes.

In this work, we investigate the visual learning problem with text descriptive features from an information-theoretic perspective. In particular, our motivation comes from two desiderata: compression and invariance (to domain shifts). The information bottleneck perspective encourages representations to compress the input as much as possible while maintaining high mutual information with the labels. On the other hand, the invariance principle favors representations that are less informative about the domains, in particular, the mutual information between the representations and the domain index should be small \citep{zhao2022fundamental,li2021learning,li2022invariant,zhao2019learning,arjovsky2019invariant,ahuja2021invariance}. Rooted in these information-theoretic principles, we propose a simple and effective method to generate classifiers based upon multiple automatically-extracted visual descriptors of each class. Our new method, SLR-AVD (Sparse Logistic Regression using Augmented Visual Descriptors), uses a language model to extract multiple potential visual features of each class, then uses $\ell_1$-regularized logistic regression to fit a sparse linear classifier on top of these visual descriptions. The key observation that supports our method is that these descriptive features retain substantial information about the true labels, yet are not informative about the domain index, making them good invariant representations of the images. Additionally, these descriptive features are better input compressors and thus can generalize better. 

% the (sparsely selected) visual descriptors are able to distinguish different classes, even if the multi-modal model is not trained on such supervision explicitly, and also provide an intuitive understanding of which features are important to which classes. 

Once the important visual descriptors are selected, we can also finetune the image encoder with the selected sparse pattern to further improve classification accuracies.  Using this procedure, SLR-AVD outperforms baselines on both in-distribution (ID) and out-of-distribution (OOD) image classification across a range of image datasets.  Specifically, SLR-AVD on ImageNet and its variations (including ImageNet-R, ImageNet V2, etc.) outperform linear probing with image features by $6.2\%$ to $10.48\%$ varying $k$-shot from $k=1$ to $k=32$. When combining SLR-AVD with WISE-FT \citep{wortsman2022robust}, on the in-distribution task, our method outperforms standard finetuning by $1.43\%$ with $1$-shot, $1.62\%$ with $2$-shot, and $1.61\%$ with $4$-shot training data. When we average over five ImageNet variations, we outperform standard finetuning by $0.88\%$ with $1$-shot, $0.73\%$ with $2$-shot, and $0.64\%$ with $4$-shot training data. 
% \anna{When combining our method with WISE-FT, our finetuning method outperforms standard finetuning by... with $4$-shot training data. When we average over five ImageNet variations (including ImageNet-R, ImageNet V2, etc), we outperform ...}

%\subsection{Notation}
\paragraph{Notation} Throughout the paper, we use $g(\cdot)$ to denote the text encoder and $f(\cdot)$ to denote the image encoder. We use $\vt$ for text tokens and $\vp$ for images. For a vector $\vv$, subscripted $\vv_i$ represents the $i$th entry. We sometimes overload the notation $\vt_c$ to represent a vector belonging to a class $c$, this should be clear from the context. We use  $\mathcal{C}$ to denote the set of classes. We use $I(X; Y)$ to denote the mutual information between a pair of random variables $(X, Y)$.
\section{Related works and motivation}
\begin{figure*}[tb]
\centering
    \centering
    \includegraphics[width=\dimexpr\linewidth-25pt\relax]
    {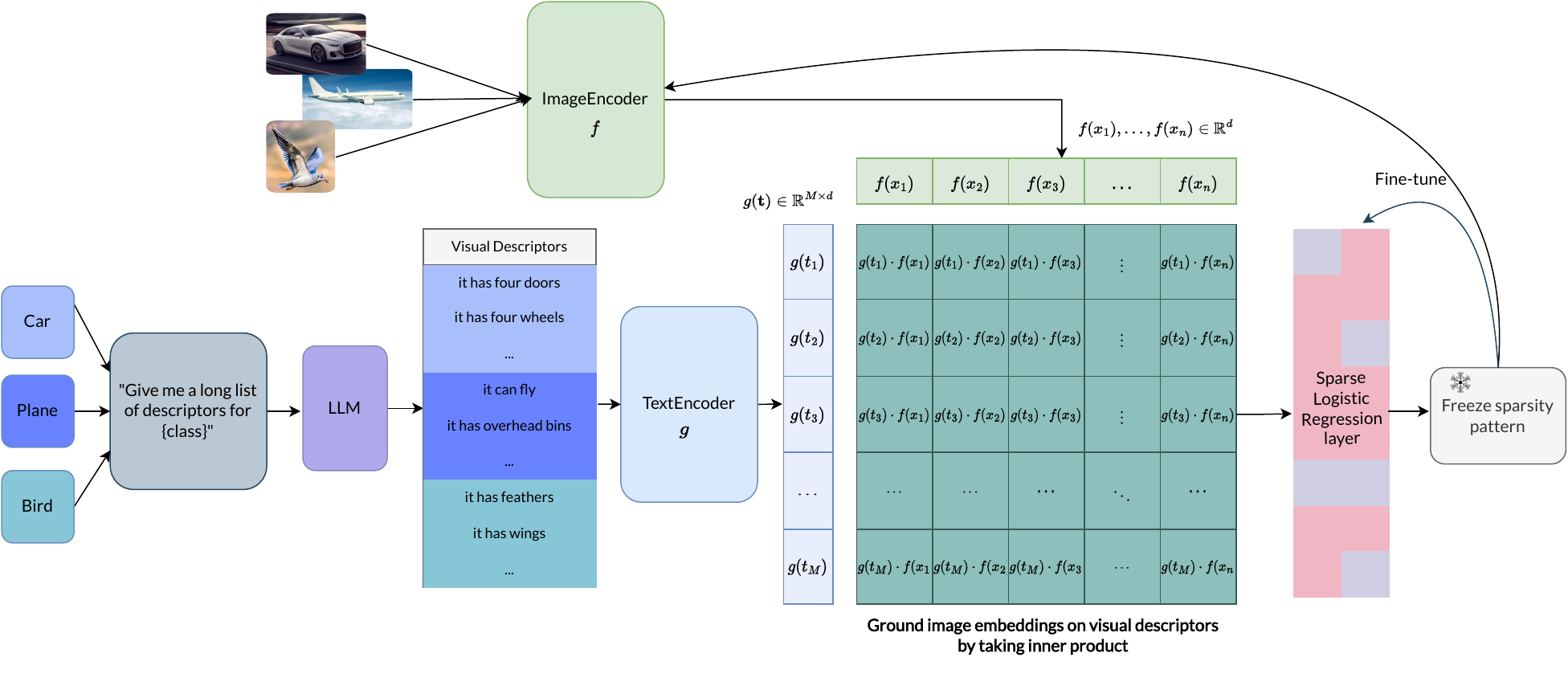}
\caption{An overview of our proposed method. We prompt GPT-3 for a list of visual descriptors for each class and encode these texts. The image embeddings are instantiated to these descriptors by taking inner products. For an image embedding in $\R^d$, this operation projects it onto a $\R^M$ dimensional space, but it may live in a submanifold. We apply sparse logistic regression over all $\R^{n\times M}$ training data for feature selection. Finally, we \textbf{freeze} the sparsity pattern and finetune both the linear layer and the image encoder to align the image features with the visual descriptors.}
\label{fig:proposed}
\end{figure*}

\subsection{Prompt tuning in VLMs}
%\paragraph
Contrastive VLMs aim to minimize the contrastive loss between matching image-text pairs. Let the image embedding be $f(\vp)\in\R^{(1+M)\times d}$, the text embedding be $g(\vt)\in\R^{(1+P)\times d}$. WLOG, let the first entry of the embeddings be the [CLS] token, denote as $g(\vt)_0$. The probability of the prediction is then represented as:
%\begin{align*}
%\begin{split}
%	p(y=c|\vp, \vt)=\frac{\exp\p{\ip{f(\vp)_0}{ g(\vt_c)_0} /\tau} } {\sum_{c'}\exp\p{\ip{f(\vp)_0}{ g(\vt_{c'})_0} /\tau}},
%\end{split}
%\end{align*}
$
	p(y=c|\vp, \vt)=\frac{\exp\p{\ip{f(\vp)_0}{ g(\vt_c)_0} /\tau} } {\sum_{c'}\exp\p{\ip{f(\vp)_0}{ g(\vt_{c'})_0} /\tau}},
$
where $\vt_c$ is the zero-shot text prompt for class $c$.
The class whose prompt has the largest inner product with the image embedding will be the zero-shot prediction. \citet{zhou2022learning} optimizes over the continuous text embedding space for the best prompts. Several follow-up works \citep{zhou2022conditional,zhu2022prompt} propose various prompt tuning methods for different task settings.  The methods that eventually use $g(\vt_c)_0$ are in essence regularized linear probing where the search space is constrained by the co-domain of $g(\cdot)_0$. \citet{chen2022prompt} uses local information of the image embedding $f_1,\ldots, f_{M+1}$ for optimizing an optimal transport distance between local image information and prompts. \citet{lu2022prompt} learns distributions over prompts for efficient adaptation to downstream recognition tasks. \citet{wen2023hard} discusses discrete prompt search in the context of text-to-image settings. 

\citet{pratt2022does} prompts LLMs for descriptions of each class and shows that these prompts can achieve better zero-shot image classification accuracy. \citet{menon2022visual} prompts LLMs to generate visual descriptors for image classification. For each class $c$, they query GPT-3 using the prompt ``\texttt{What are useful features for distinguishing a \{$c$\} in a photo}?''. A score is estimated for $c$ given an image $\vp$:
$
	s(c, \vp) = \frac{1}{|D(c)|}\sum_{\vt\in D(c)}\phi(\vt, \vp),
$ 
where $D(c)$ is the set of descriptors for $c$, and $\phi(\vt, \vp)=\ip{f(\vp)_0}{g(\vt)_0}$ is the inner product between the image and text embeddings. They show this average ensemble can outperform zero-shot classifiers while maintaining interpretability. 

Similar to what we propose, LaBo \citep{yang2022language} also considers per-class level descriptions in the few-shot setting. A key difference is that they perform a per-class level description filtering through submodular optimization, and they apply softmax to a linear weight $\sigma(\mW)$ to ensemble the selected features. On the other hand, we directly select features using sparse logistic regression. Our approach immediately gives both the important features and the coefficients and is statistically optimal under certain sparsity assumptions. One of the potential drawbacks of LaBo is their visual descriptions are filtered per-class level, which can hinder feature sharing between classes.
% This partly explains why on ImageNet, LaBo starts to underperform linear probing when $k>=4$ (see Figure 3 in their paper). 
LaBo uses $\sigma(\mW)$ in order to gain probabilistic interpretations of the features, while our emphasis on robustness only requires $\mW$ to be sparse.

\subsection{Robust fine-tuning of zero-shot models}
There are numerous works that study robust finetuning of zero-shot models \citep{goyal2022finetune,kumar2022fine,wortsman2022robust}. In this work, we adopt the weight interpolation method WISE-FT to improve the OOD test accuracy \citep{wortsman2022robust}.
In general, let $\bm\Phi$ refer to any set of weights in the network (just the linear layer, linear layer + image encoder, etc). Let the finetuned weight be 
%Let the finetuned weight (could be just the final linear layer or the whole network, depending on the context) be 
$\bm\Phi\_{learned}$ and let the zero-shot predictor be $\bm\Phi\_{zs}$. \citet{wortsman2022robust} observes that while $\bm\Phi\_{learned}$ performs better than $\bm\Phi\_{zs}$ on ID tasks, it is worse at OOD tasks. Hence they propose to interpolate the two sets of weights as $\alpha \bm\Phi\_{learned} + (1-\alpha) \bm\Phi\_{zs}$. This surprisingly simple weight ensemble helps both in-distribution and out-of-distribution tasks. This method also naturally applies to linear probing by simply freezing the CLIP encoder throughout, and only training and interpolating the linear head.

\subsection{Compression and Invariant Representation}
The term ``compression'' has been given various meanings under different contexts. \citet{arora2018stronger} derived a PAC bound where generalization depends on the compression of the model parameters; \citet{moran2016sample} developed a sample compression scheme where both the features and labels are compressed; information bottleneck  \citep{tishby2015deep} proposed to learn representations $Z$ that ``compresses'' the inputs $X$ by minimizing $I(X;Z)$ subject to some constraints. \citet{blier2018description,blum2003pac} discussed label compression in terms of model description length. In this work, we use the term to represent input compression (as in the information bottleneck), such that the features contain little information about the inputs. From a PAC-learning perspective, a better input compression will lead to a smaller generalization error \citep{shwartz2018representation,galloway2022bounding}, motivating our use of text descriptive features.
%One should note that any label compression scheme is essentially an estimation of the mutual information, since $H(Y)-\E[L(Y|X)]\leq H(Y)-H(Y|X)=I(Y;X)$, where $L(Y|X)$ is the codelength of $Y$ given $X$. However, since $L(\cdot|\cdot)$ is restricted by its model class and finite sample size, it is unclear how $I(Y;X)$ will affect the codelength \textit{algorithmically}. We will later show that the text descriptive features will often lead to shorter codelength, despite a smaller mutual information with the inputs, compared to the image embeddings.
A complementary idea from information theory is the invariance principle. The idea is that we want to learn representations that are very informative about the labels, but not so about the domain information. Mathematically, the principle encourages $\max_Z I(Y;Z)-\lambda I(Z; A)$ where $A$ is the domain information \citep{zhao2022fundamental}. While it is understood that invariance by itself is insufficient for OOD generalization \citep{ahuja2021invariance,rosenfeld2020risks}, algorithms based on the invariance principle still achieve competitive results on several OOD benchmarks \citep{koh2021wilds}.

\section{Proposed method}\label{sec:propose}

In this section, we present our proposed method, SLR-AVD, summarized in \cref{fig:proposed}. We will discuss how to generate features, select a sparse set of useful descriptions, and finally, how to align the encoder in detail. We will also state how the proposed method aligns with information-theoretic principles.

\subsection{Generating visual descriptors}\label{subsec:genvd}
To generate the visual descriptors for ImageNet and its variations, we first use the following prompt to query GPT-3: ``\texttt{Give me a long list of descriptions for \{\}:}''.

GPT-3 is quite sensitive to format instruction. Using the prompt ``Give me a list'' always leads to a list format, making it straightforward to select the useful text with regular expressions. 
% We also did not explicitly ask GPT-3 for visual features. As we have shown in \cref{subsec:observation}\zico{fix this reference}xt, even abstract text can be used for classification. 
Following the method in \citet{menon2022visual}, we condition these descriptors on the class name, using texts of the form ``\texttt{$\{c\}$ which has \{$\vt_c^i$\}}'' for each class $c$ and the $i$th descriptor. For each class $c$, we gather $M_c$ descriptors from GPT-3.

Furthermore, for each class, there exists a set of hand-crafted prompt templates like ``\texttt{a photo of \{\}}'' or ``\texttt{an art of \{\}}''. If there are $T$ total number of such templates, using the class name $c$, we can generate $T$ total prompt embeddings for each class. We take the average of these prompt embeddings \textit{in addition to} the aforementioned visual descriptors, leading to $M_c+1$ number of prompts for each class. For simplicity, we will refer to the GPT-3 generated text features as \textit{visual descriptors (VD)}, the templates with class names as \textit{class prompts} (CP), and the union as \textit{augmented visual descriptors (AVD)}. We will also refer to their \textit{embeddings} using the same names, which should be clear from the context.   

Denote $M=\sum_{c\in\cC}M_c$ where $\cC$ is the set of all classes. The visual descriptors, class prompts, and augmented visual descriptors can be encoded into three matrices $\mU\_{vd}\in\R^{M\times d},\mU\_{cp}\in\R^{|\cC|\times d},\mU\_{avd}\in\R^{(M+|\cC|)\times d}$. Given an image embedding $\vz:=f(\vp)_0\in\R^d$, these three matrices respectively created three sets of new features $\vh\_{vd}=\mU\_{vd}\vz$, $\vh\_{cp}=\mU\_{cp}\vz$, and $\vh\_{avd}=\mU\_{avd}\vz$. Notice that all three $\mU$ matrices are fixed and never trained. We call the action of inner product $\ip{\mU}{\cdot}$ as ``instantiating''. We will also refer to the instantiated features $\vh$ as the (text/language) descriptive features. Given $\vh$, we can learn three matrices $\mW\_{vd}\in\R^{|\cC|\times M},\mW\_{cp}\in\R^{|\cC|\times |\cC|},\mW\_{avd}\in\R^{|\cC|\times (M+|\cC|)}$. 

Setting 
%$
%\mW\_{vd}=\operatorname{blkdiag}\p{\p{\underbrace{\frac{1}{|M_c|},\ldots, \frac{1}{|M_c|}}_\text{$|M_c|$ copies}}_{c\in\cC}},
%$ then 
$
\mW\_{vd}=\operatorname{blkdiag}\big((\underbrace{\frac{1}{|M_c|},\ldots, \frac{1}{|M_c|}}_\text{$|M_c|$ copies})_{c\in\cC}\big),
$ 
then
$\mW\_{vd}\mU\_{vd}$ leads to the average ensemble in \citet{menon2022visual}. Setting $\mW\_{cp}=I_{|\cC|\times|\cC|}$, we get back the zero-shot classifier $\mW\_{zs}=\mW\_{cp}\mU\_{cp}$. One can naturally merge $\mW\_{vd}$ and $\mW\_{cp}$ into $\mW\_{avd}=[\mW\_{vd}, \mW\_{cp}]$, which we use in our proposed method. We note that these three $\mW$ matrices can all serve as zero-shot classifiers.

\subsection{Learning sparse ensemble and aligning the image encoder}

The previously defined matrix $\mU\_{avd}$ can be viewed as a linear projection of the image embedding onto a $M+|\cC|$ dimensional semantic space. While this space has a high ambient dimension, the projected embeddings live in a low-dimensional manifold that has rank less than or equal to that of the image embedding space. By enforcing a sparsity constraint on $\mW\_{avd}$, we can select the most important dimensions among $\vh\_{avd}$.  We demonstrate that the selected subspace is also robust to natural distribution shifts. Intuitively, we imagine that the large distribution shift in the image embedding space only corresponds to a small shift in the semantic space, since the semantics of images should be invariant. We will later demonstrate with mutual information estimations. Further investigation on the property of the semantic space is left to future works.

With a fixed $\mU\_{avd}$, we learn $\mW\_{avd}$ with $\ell_1$ regularization $\norm{\mW\_{avd}}_1$. Not only does sparse logistic regression select the important features, but it actually also finds the intuitive features. For example, on CIFAR-10, we demonstrate that the selected features are usually the ones that actually describe that class: for each class, we pick the three features with the largest coefficients, and show that the properly descriptive class features are chosen most often; the results are listed in \cref{table:cifar_features} in the appendix.  
%
%
%\subsection{Aligning the image encoder}
After obtaining a sparse $\widehat\mW\_{avd}$, we fix $\mU\_{avd}$ and the \textit{sparsity pattern} of $\widehat\mW\_{avd}$, and finetune both the image encoder $f$, as well as the entries in $\widehat\mW\_{avd}$. This process aligns with LP-FT \citep{kumar2022fine}, which has some theoretical justification for its robustness. 
\subsection{Text descriptive features are compressive and invariant}
Beyond the improvement in performance alone, however, the core of our method relies on the empirical evidence that text descriptive features have many benefits from an information-theoretic perspective. Specifically, we show here that the text descriptive features form more \emph{invariant} and more \emph{compressive} representations of the data than the naive image encoder features.  This motivates their use, especially under distribution shift, where we see them outperform the alternatives.

% \paragraph{Feature invariance}
We base our investigation upon two notions: the invariance principle and the information bottleneck. First, the invariance principle from causality \citep{pearl1995causal} states that the predictors should only rely on the causes of the labels rather than the spurious features. Following this principle, several mutual information (MI) based OOD generalization works \citep{arjovsky2019invariant,zhao2022fundamental,li2021learning,li2022invariant,zhao2019learning,feng2021provable,ahuja2021invariance} propose that a good feature representation $Z$ would have high mutual information with the label, $I(Z;Y)$, but low MI with the domain index, $I(Z;A)$, so as not to leak information about the domain itself.  Closely related is the information bottleneck, which similarly states that a good representation will again have high MI with the label, but low MI with the input $I(Z;X)$.  In recent years, several works have suggested that combining the invariance principle with the information bottleneck can lead to practical and provably strong OOD generalization \citep{ahuja2021invariance,li2022invariant}.

%To recapitulate, we investigate visual learning with text descriptive features from three aspects: IB, invariance, and label compression. \anna{To summarize, we investigate visual learning with text descriptive features from three angles: IB, invariance, and label compression.}
%The invariance principle is formulated as $\max_Z I(Z; Y)-\lambda I(Z;A)$. An invariant feature mapping is expected to have small $I(Z;A)$, but since $A$ may very well contain information about $Y$, there is then an intrinsic trade-off between the two mutual information terms. IB proposes to minimize the objective $\min_Z I(Z; X)-\beta I(Z; Y)$, and in recent years people have suggested that combining invariance and IB can lead to practically and provably good OOD generalization 

We demonstrate that the text descriptive features essentially obey both the tenets of the invariance principle and the information bottleneck: the extracted text features $H$ have high MI with the labels, but substantially lower MI with both the domain index and the input itself.
% \anna{We will demonstrate that the descriptive features leak much less information about the domain index than the image embeddings, and therefore they align more with the invariance principle in OOD generalization.} 
The features of our framework correspond to the following Markov chain:
%\begin{align}\label{eq:infoflow}
%\begin{split}
%	\vp \xrightarrow{f(\cdot)_0} \vz \xrightarrow{\mU} \vh \xrightarrow{\mW} \hat\vy,
%\end{split}
%\end{align}
%$
%	\vp \xrightarrow{f(\cdot)_0} \vz \xrightarrow{\mU} \vh \xrightarrow{\mW} \hat\vy,
%$
\begin{equation}
	Y\rightarrow X \xrightarrow{f(\cdot)_0} Z \xrightarrow{\mU} H \xrightarrow{\mW} \hat Y,
\end{equation}
where $\vy\sim Y$, $\vp\sim X, \vz\sim Z$, $\vh\sim H$, $\hat \vy\sim \hat Y$ corresponds to realizations of the truth labels, the input images, the image embeddings, the text descriptive features, and the predictions (the capital letters are random variables) respectively. % is the input image, $\vz = f(\vp)_0$ is the (CLIP) image embedding of $\vp$, $\vh=\mU\vz$ is the grounded feature, and $\hat\vy=\argmax_i\{(\mW\vh)_1,\ldots, (\mW\vh)_{|\cC|}\}$ is the prediction. 
 Here $\mW, \mU, \vh$ and $H$ can be subscribed by $\operatorname{avd},\operatorname{vd},\operatorname{cp}$ as in \cref{sec:propose}. We will use $A$ for the domain index.
%
%\begin{wrapfigure}{rlio}{0.45\linewidth}
%\centering
%        \includegraphics[width=0.99\linewidth]{figures/mi_y_domain_smile.pdf}
%        \includegraphics[width=0.96\linewidth]{figures/mi_x_smile.pdf}
%    \caption{The MI estimations at interest. \textbf{Top:} the MI between a different set of features and the labels or the domain indices. \textbf{Bottom}: the MI between a different set of features and the input images.}
%%    \vspace{-2em}
%\label{fig:mis}
%\end{wrapfigure}

By the Data Processing Inequality (DPI, \citet{cover1999elements}), we immediately have that $I(X;Y) \geq I(Z;Y) \geq I(H; Y)$. Additionally, however, we also observe for the text descriptive features $I(H;Y)$ is nearly as large as $I(Z;Y)$ (i.e., there is not much decrease in the information about the label), but $I(H;A)$ and $I(H;X)$ are substantially lower than $I(Z;A)$ and $I(Z;X)$ (i.e, the text descriptive features leak much less information about the label and the input).
%Although seemingly a decreased MI can lead to worse performance \anna{Do you want to say degenerate performance or just decreased/worsened performance}, we find that $I(H; A)$ is significantly smaller than $I(Z; A)$. This phenomenon explains why descriptive features leads to a better OOD performance from the invariance principle. \anna{I don't exactly follow here. A is not in the Markov chain but you didn't transition to how this last claim relates to DPI.}

\begin{figure}[tb]
 	\begin{subfigure}[t]{0.44\textwidth}
 	  \includegraphics[width=\textwidth]{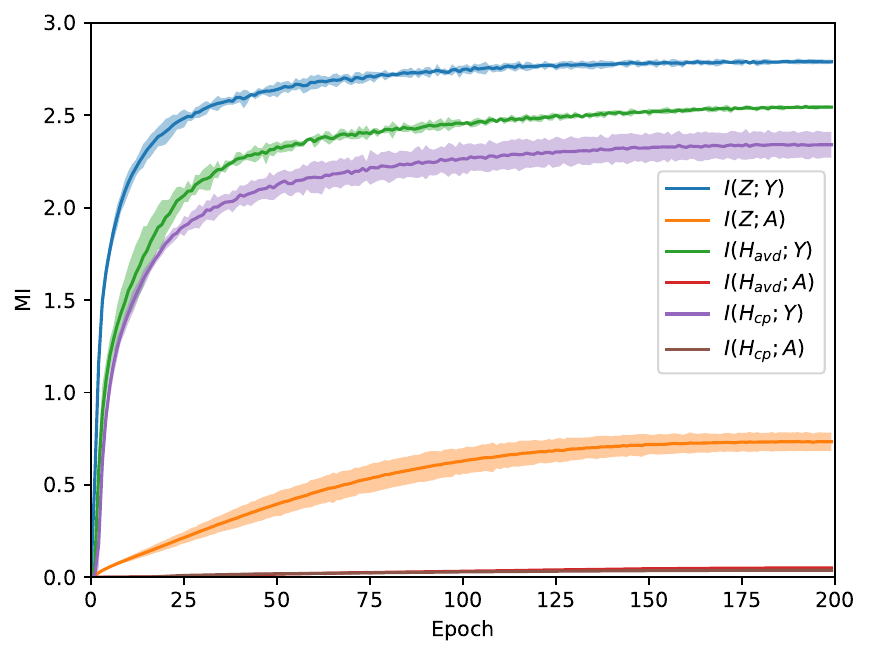}
 %	  \caption{The rectangle is a complicated geometrical figure that has 4 edges and 4 vertices while a star is an even more complex geometrical figure}
 	\end{subfigure}
 	\hfill
 	\begin{subfigure}[t]{0.44\textwidth}
 	  \includegraphics[width=\textwidth]{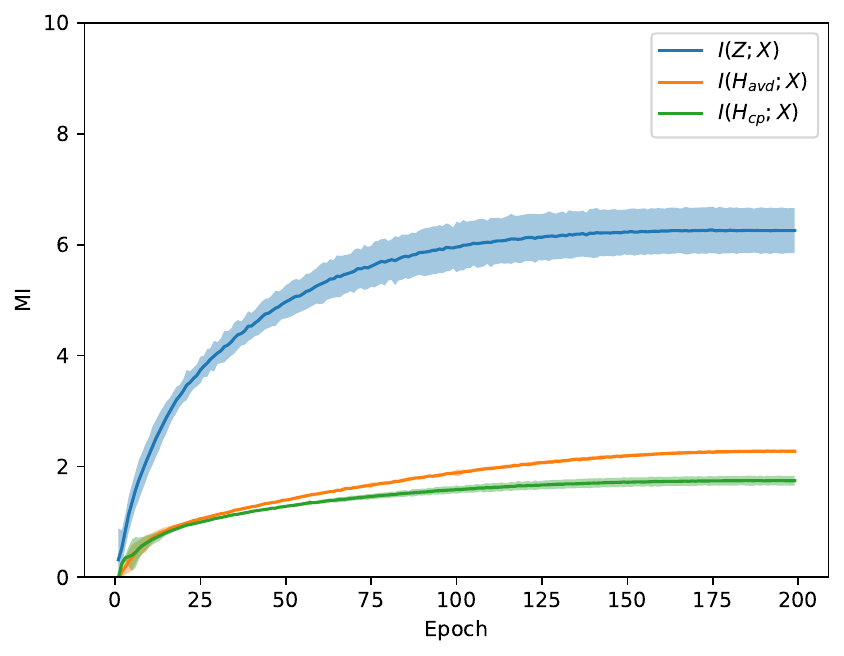}
 %	  \caption{A simple circle and a spiral}
 	\end{subfigure}
 \caption{The MI estimations at interest. The estimator is variational and we include the whole optimization trajectory. \textbf{Left:} the MI between a different set of features and the labels or the domain indices. \textbf{Right}: the MI between a different set of features and the input images.}
 \label{fig:mis}
 \end{figure}

To assess this, we conduct numerical evaluations on CIFAR-10 \citep{krizhevsky2009learning}, CIFAR-10.1 \citep{recht2018cifar}, and CIFAR-10.2 \citep{lu2020harder}. We index these three datasets, denoting the index random variable as $A$. We compute the image embedding $\vz$ and the instantiated descriptive feature $\vh$ for every image in these three test sets. To estimate mutual information, we use the SMILE estimator \citep{song2019understanding}. The numerical estimation is presented in \cref{fig:mis}. MI is estimated for two sets of text descriptive features: $\vh_{\mathrm{cp}}\sim H_{\mathrm{cp}}$ and $\vh_{\mathrm{avd}}\sim H_{\mathrm{avd}}$. Importantly, $H_{\mathrm{cp}}$ should be viewed as a post-processing of $H_{\mathrm{avd}}$.
Intuitively, we see that $I(Z; Y)>I(H_{\mathrm{avd}}; Y)>I(H_{\mathrm{cp}}; Y)$ by DPI. We also see that $I(Z; A)>I(H_{\mathrm{avd}}; A)>I(H_{\mathrm{cp}}; A)$, which suggests that the text descriptive features $\vh$ are much more invariant to the distribution shift. The noticeable gap between $I(H_{\mathrm{avd}}, Y)$ and $I(H_{\mathrm{cp}}, Y)$ explains why it is beneficial to work with text descriptive features beyond vanilla zero-shot classification. 

From the information bottleneck perspective, \Cref{fig:mis} also presents that $I(X; H_{\mathrm{avd}})< I(X; Z)$ by a large margin, we can then interpret $H_{\mathrm{avd}}$ as a ``better'' compression of the input image $X$, in the sense that it preserves only information in $X$ that is helpful for predicting $Y$. Of course, this also means that one cannot reconstruct $X$ from $H_{\mathrm{avd}}$ better than from $Z$, although this is an orthogonal goal to ours. Typically better input compressions lead to smaller generalization error. Under mild conditions one can bound the generalization error of feature $Z$ with probability at least $1-\delta$:
$
    \mathrm{GenErr}\leq\sqrt{\frac{2^{I(X;Z)}+\log(2/\delta)}{n}},
$
where $n$ is the number of training samples \citep{shwartz2018representation}. Intuitively, if the features have small MI with the inputs, then the perturbation in the input space cannot perturb the features too much, hence constraining the expressiveness of the features. Since $I(H_{\mathrm{avd}}; X)$ is significantly smaller than $I(Z; X)$, we can expect a more predictable test performance (compared to the training performance). On the other hand, high $I(H_{\mathrm{avd}}; Y)$ makes sure that the accuracy will not be too low. The synergy of the two notions elucidates the superiority of AVD in the few-shot setting.

%\begin{table}[tb]
% \begin{tabular}{c|c|c|c|c|c|c} 
% \toprule
%\multirow{2}{*}{Code} & \multicolumn{2}{c}{CIFAR-10} & \multicolumn{2}{c}{CIFAR-10.1} & \multicolumn{2}{c}{CIFAR-10.2}\\
% & Codelength & Comp ratio & Codelength & Comp ratio & Codelength & Comp ratio\\
% \midrule
% Uniform & $33219$ & $1.0$ & $6644$ & $1.0$ & $6644$ & $1.0$ \\
% $H_{avd}$ & $3228$ & $0.0972$ & $1329$ & $0.2001$ & $1474$ & $0.2122$\\
% $Z$ & $6438$ & $0.1938$ & $3094$ & $0.4657$ & $1491$ & $0.2244$\\
%\bottomrule
% \end{tabular}
% \caption{\textcolor{red}{This needs a caption?}}
%\end{table}

\section{Experiment}

% \begin{figure*}[t]
%      \centering
%      \begin{subfigure}[b]{0.28\textwidth}
%          \centering
%          \includegraphics[width=\textwidth]{figures/fs_img_vs_text_ind.pdf}
%      \end{subfigure}
%      \hfill
%      \begin{subfigure}[b]{0.28\textwidth}
%          \centering
%          \includegraphics[width=\textwidth]{figures/fs_img_vs_text_r.pdf}

%      \end{subfigure}
%      \hfill
%      \begin{subfigure}[b]{0.28\textwidth}
%          \centering
%          \includegraphics[width=\textwidth]{figures/fs_img_vs_text_a.pdf}

%      \end{subfigure}
     
%      \begin{subfigure}[b]{0.28\textwidth}
%          \centering
%          \includegraphics[width=\textwidth]{figures/fs_img_vs_text_v2.pdf}

%      \end{subfigure}
%      \hfill
%      \begin{subfigure}[b]{0.28\textwidth}
%          \centering
%          \includegraphics[width=\textwidth]{figures/fs_img_vs_text_sketch.pdf}

%      \end{subfigure}
%      \hfill
%      \begin{subfigure}[b]{0.28\textwidth}
%          \centering
%          \includegraphics[width=\textwidth]{figures/fs_img_vs_text_object.pdf}

%      \end{subfigure}
%         \caption{Few-shot experiments compare LP vs SLR-AVD. In each subfigure, the y-axis represents the number of shots per class. Here we consider shot in $[1,2,4,8,16,32]$ per class. SLR-AVD is more sample efficient in the in-distribution reference and is also much more robust to several distribution shifts.}
%         \label{fig:fs_comparison}
% \end{figure*}

\begin{figure}[tb]
    \centering
    \includegraphics[width=\textwidth]{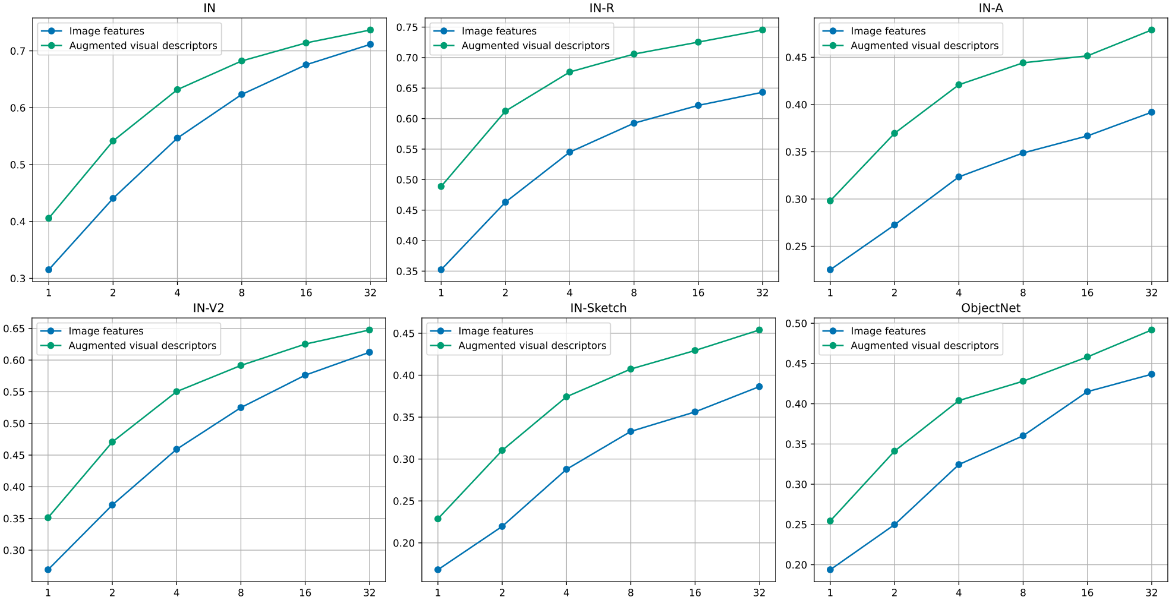}
    \caption{Few-shot experiments compare LP vs SLR-AVD. In each subfigure, the y-axis represents the number of shots per class. Here we consider shot in $\{1,2,4,8,16,32\}$ shots per class. SLR-AVD is more sample efficient in the in-distribution reference and is also much more robust to several distribution shifts.}
    \label{fig:fs_comparison}
\end{figure}

Throughout the experiments, we focus on the few-shot setting. We test our method on ImageNet,  ImageNet-R, ImageNet-V2, ImageNet-A, ImageNet-Sketch, and ObjectNet \citep{deng2009imagenet,hendrycks2021many,hendrycks2021natural,recht2019imagenet,wang2019learning,barbu2019objectnet}, demonstrating the superiority of the sparsely learned visual descriptors ensemble. By default, we use the ViT-B/16 model unless otherwise specified. The hand-crafted templates for ImageNet classes contain a set of seven prompts suggested in \href{https://github.com/openai/CLIP}{https://github.com/openai/CLIP}:
\begin{enumerate*}[series = tobecont, itemjoin =\quad]
	\item ``\texttt{itap of a \{\}.}''
	\item ``\texttt{a bad photo of the  \{\}.}''
	\item ``\texttt{a origami \{\}.}''
	\item ``\texttt{a photo of the large \{\}.}'
	\item ``\texttt{a \{\} in a video game.}''
    \item ``\texttt{art of the \{\}.}''
	\item ``\texttt{a photo of the small \{\}.}''
\end{enumerate*} 
This set usually outperforms the original $80$ templates in \citet{radford2021learning}.

\begin{table}[tb]
\begin{minipage}[t]{0.4\textwidth}\vspace{0pt}\centering
 \caption{Accuracies of zero-shot, visual descriptors, and augmented visual descriptors on ImageNet and its variations. ZS-AVD outperforms all baselines across different datasets.}
\resizebox{\columnwidth}{!}{
 \begin{tabular}{c|ccc} 
 \toprule
 & ZS & ZS-VD & ZS-AVD \\
 \midrule
 IN & 68.78 & 65.89 & \bf 69.52\\ 
 IN-V2 & 62.23 & 59.19 & \bf 62.97\\ 
 IN-R & 77.72 & 72.75 & \bf 77.85\\ 
 IN-A & 50.64 & 46.11 &\bf  50.87\\ 
 IN-Sketch & 48.38 & 44.84 & \bf 48.91\\ 
 ObjectNet & 54.31 & 49.60 &\bf  54.58\\ 
 \bottomrule
 \end{tabular}
 }
 \label{table:zs_comparison}
\end{minipage}
  \hfill
\begin{minipage}[t]{0.55\textwidth}\vspace{0pt}\centering
 \captionof{table}{WISE-FT vs. WISE-SLR accuracies on ImageNet and its variations with optimal $\alpha$.}
\resizebox{\columnwidth}{!}{
 \begin{tabular}{c|c|c|c|c|c|c} 
 \toprule
 Shot&  \multicolumn{2}{c|}{$k=1$} & \multicolumn{2}{c|}{$k=2$} & \multicolumn{2}{c}{$k=4$}\\
 \midrule
Method& FT & SLR & FT & SLR &FT & SLR\\
\midrule
IN & 68.88 & 70.31 & 69.59 & 71.21 & 70.48 & 72.09\\
\midrule
Average $\uparrow$ &  \multicolumn{2}{c|}{1.43} &  \multicolumn{2}{c|}{1.62} &  \multicolumn{2}{c}{1.61}\\
\midrule
IN-R & 77.82 & 78.29 & 78.13 & 78.53 & 78.32 & 78.59\\
IN-A & 50.09 & 51.29 & 50.43 & 51.51 & 52.11 & 52.64 \\
IN-V2 & 62.32 & 63.74 & 63.07 & 64.37 & 63.50 & 65.30\\
IN-Sketch & 48.45 & 49.35 & 48.75 & 49.63 & 48.99 & 49.92\\
ObjectNet & 54.52 & 54.94 & 55.01 & 54.99 & 55.77 & 55.41\\
\midrule
Average $\uparrow$ &  \multicolumn{2}{c|}{0.88} &  \multicolumn{2}{c|}{0.73} &  \multicolumn{2}{c}{0.64}\\
\bottomrule
 \end{tabular}
}

 \label{table:optimal_alpha_ft}
\end{minipage}
  \end{table}

\begin{table*}[t]
\centering
 \caption{Accuracies on ImageNet and its variation. We compare LP vs SLR-AVD.}

\resizebox{\columnwidth}{!}{
 \begin{tabular}{c|cc|cc|cc|cc|cc|cc} 
 \midrule
 Shots & \multicolumn{2}{c|}{$k=1$}  &  \multicolumn{2}{c|}{$k=2$} & \multicolumn{2}{c|}{$k=4$}&  \multicolumn{2}{c|}{$k=8$} &  \multicolumn{2}{c|}{$k=16$} &  \multicolumn{2}{c}{$k=32$} \\
 \midrule
 Methods & LP & AVD & LP & AVD  & LP & AVD &  LP & AVD &  LP & AVD &  LP & AVD \\
 \midrule
IN & 31.51 & \bf 40.56 & 44.06 & \bf 54.16 & 54.66 & \bf 63.19 & 62.33 & \bf 68.23 & 67.55 & \bf 71.40 & 71.15 & \bf 73.67 \\
IN-R & 35.23 & \bf 48.88 & 46.30 & \bf 61.23 & 54.50 & \bf 67.64 & 59.25 & \bf 70.58 & 62.16 & \bf 72.54 & 64.32 & \bf 74.53 \\
IN-A & 22.52 & \bf 29.81 & 27.26 & \bf 36.96 & 32.34 & \bf 42.09 & 34.88 & \bf 44.41 & 36.68 & \bf 45.15 & 39.19 & \bf 47.89 \\
IN-V2 & 26.91 & \bf 35.12 & 37.13 & \bf 47.07 & 45.92 & \bf 55.02 & 52.50 & \bf 59.15 & 57.62 & \bf 62.52 & 61.23 & \bf 64.75 \\
IN-Sketch & 16.80 & \bf 22.87 & 21.96 & \bf 31.03 & 28.77 & \bf 37.43 & 33.29 & \bf 40.73 & 35.62 & \bf 42.94 & 38.64 & \bf 45.39 \\
ObjectNet & 19.38 & \bf 25.43 & 24.98 & \bf 34.11 & 32.44 & \bf 40.39 & 36.02 & \bf 42.80 & 41.50 & \bf 45.82 & 43.67 & \bf 49.17 \\
\midrule
Average $\uparrow$ & \multicolumn{2}{c|}{8.39} & \multicolumn{2}{c|}{10.48} & \multicolumn{2}{c|}{9.52} & \multicolumn{2}{c|}{7.94} & \multicolumn{2}{c|}{6.54} & \multicolumn{2}{c|}{6.20} \\
 \bottomrule
 \end{tabular}
 }
 \label{table:fs_comparison}
\end{table*}

For simplicity, we will use the following acronyms for different methods and datasets. We defer the hyperparameter discussions to the appendix.

\noindent\textbf{ZS:} Zero-shot classification using text embeddings of hand-crafted prompts ensembles. 
\noindent\textbf{ZS-VD, ZS-AVD:} Zero-shot classification using visual descriptor and augmented visual descriptors, respectively.
\noindent\textbf{LP:} Linear probing using image embeddings.
\noindent\textbf{SLR-AVD:} Sparse logistic regression using AVDs. 
\noindent\textbf{FT:} Finetuning the image encoder and classification head.
\noindent\textbf{SLR-FT-AVD:} Sparse logistic regression with AVD, and then finetune the linear head plus the image encoder with frozen sparsity patterns.
\noindent\textbf{WISE-FT:} Weight ensemble using ZS and FT.
\noindent\textbf{WISE-SLR:} Weight ensemble using SLR-FT-AVD and ZS-AVD. 
\noindent\textbf{IN:} ImageNet.
\noindent\textbf{IN-R:} ImageNet-R.
\noindent\textbf{IN-A:} ImageNet-A.
\noindent\textbf{IN-V2:} ImageNetV2.
\noindent\textbf{IN-Sketch:} ImageNet-Sketch.
%
%\paragraph{Hyperparameter} For ImageNet and its variations, we fix a set of 6804 augmented visual descriptors. The hyperparameters are swept over disjoint training and validation sets of size $20$ per class for LP and SLR-AVD. For $\ell_1$ regularization, its non-smoothness makes it notoriously hard for auto-differentiation. To circumvent the smoothness issue, we apply the GPU implementation \citep{wong2021leveraging} of a variance-reduction proximal gradient method SAGA \citep{defazio2014saga}.
%We adopt the \textit{regularization path} approach, in which the solver optimizes over $100$ regularization strengths $\lambda_1>\lambda_2\cdots >\lambda_{100}$. Here we set $\lambda_1$ to be the strength that returns a model that uses none of the features, and $\lambda_{100}=0.1\times\lambda_1$. For LP, we always use $\ell_2$ regularization, we use L-BFGS implemented by scikit-learn, and search for the regularization strength over $100$ grids between $0.5$ and $6$. All the $\lambda$s are evenly spread in the log-space\footnote{In python numpy.logspace(math.log10($\lambda_1$), math.log10($\lambda_{100}$), 100)}. For FT and SLR-FT-AVD, we select hyperparameters using a training and validation set of size $4$ per class. The batch size is fixed to be $512$ and the number of epochs is fixed to be $10$. We always optimize with AdamW, and choose a cosine rate scheduler with warm-ups. We randomly select learning rate in $[1e-8, 3e-5]$, weight decay in $[0.1, 0.12]$, and warm up steps in $\{0, 50, 500\}$, for $20$ trials. The chosen parameters are then fixed throughout all experiments. 

\subsection{Zero-shot with AVDs}
As mentioned in \cref{subsec:genvd}, we can easily establish zero-shot matrices with AVDs. We set $\mW\_{vd}$ to be the aforementioned block diagonal form, $\mW\_{cp}$ to be an identity matrix. We merge them into $\mW\_{avd}=[\mW\_{vd}, \gamma\mW\_{cp}]$.
 Their performances are compared in \cref{table:zs_comparison}. ZS-AVD outperforms every zero-shot baseline on all ImageNet variations. We find that simply using VD usually underperforms ZS, indicating that the class names are probably one of the strongest prompts. This observation is intuitive as during contrastive training, the class name itself is likely to show up in the caption the most often, compared to other visual descriptors. 
 One can certainly try to improve ZS-VD results by more carefully prompting GPT-3, or gathering descriptors from different data sources/search engines. \citet{pratt2022does,yang2022language,menon2022visual} have studied the quality of descriptors across different datasets and hyperparameters (e.g. temperature for sampling, etc) settings. Here, we do not further pursue this direction. Instead, we utilize our observation that simply using the merged prompts $\mW\_{avd}$ already surpasses the best zero-shot classifier. Notice here we have a parameter $\gamma$ that decides how much we weight the zero-shot model. Empirically we find that setting $\gamma=5$ is sufficient for all datasets. We conduct small-scale experiments on CIFAR-10 and its variations to further investigate the influence of difference choice of $\gamma$, the GPT prompts, and the GPT sampling hyperparameters. We find these choices typically do not lead to significant deviations in test accuracies unless the generated visual descriptors are too repetitive, see the appendix for details.

\subsection{Comparison to linear probing}

We compare SLR-AVD to LP with $\{1,2,4,8,16,32\}$ shots per class. Each experiment is conducted 3 times with independent random seeds. We report the averaged test accuracy on ImageNet and its distribution shift variations, see \cref{fig:fs_comparison} for details. Our proposed method outperforms linear probing on all tasks. Detailed accuracies are presented in \cref{table:fs_comparison}. In a nutshell, our method outperforms linear probing by $8.39\%$, $10.48\%$, $9.52\%$, $7.94\%$, $6.54\%$, $6.20\%$ on $k=1,2,4,8,16,32$ respectively.

%\begin{figure}
%     \centering
%     \begin{subfigure}[b]{0.3\textwidth}
%         \centering
%         \includegraphics[width=\textwidth]{figures/fs_img_vs_text_avg.pdf}
%     \end{subfigure}
%
%        \caption{Few-shot experiments compare learning with augmented visual descriptors (AVD) vs linear probing. In each subfigure, the y-axis represents the number of shots per class. Here we consider shot in $[1,2,4,8,16,32]$ per class. The accuracies are averaged over all six ImageNet and its variation datasets, and over 3 runs. Learning with AVD is significantly more sample efficient in both in-distribution and out-of-distribution tasks.}
%        \label{fig:fs_comparison}
%\end{figure}

\begin{figure}[tb]
    \centering
    \includegraphics[width=\textwidth]{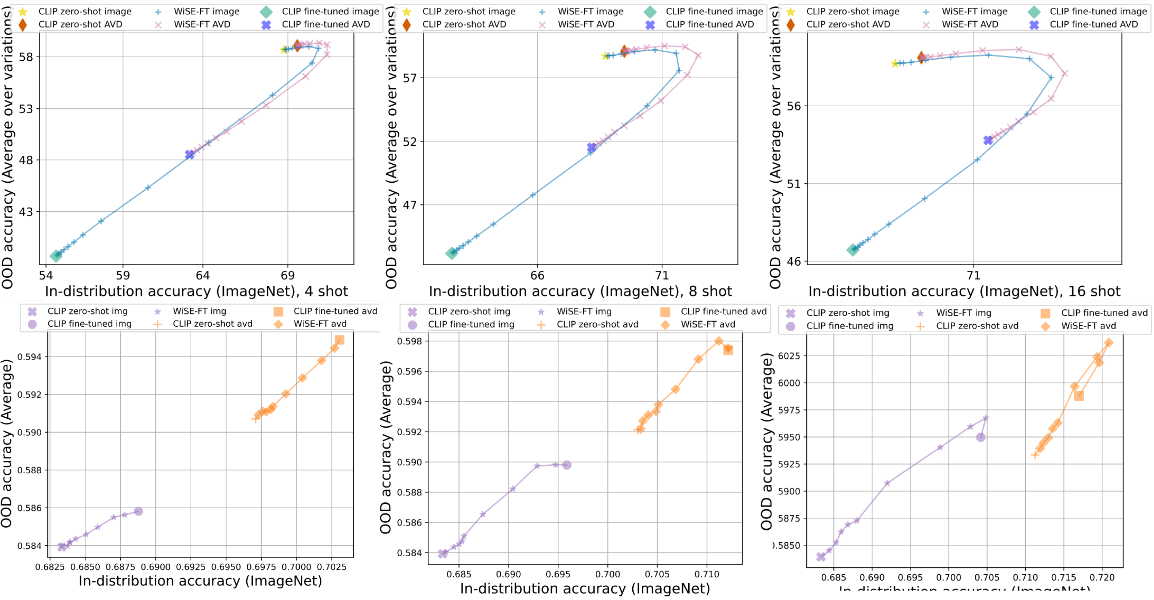}
    \caption{\textbf{Top: }ID-OOD accuracy curve of WISE-FT+LP vs WISE-FT+SLR-AVD. ID is tested on ImageNet, and OOD is averaged over $5$ ImageNet variations. Experiments with $[4,8,16]$-shots are presented. Each accuracy is averaged over $3$ runs. We can see that our proposed method overwhelms LP in all cases. \textbf{Bottom:} ID-OOD accuracy curve of WISE-FT vs WISE-SLR. ID is tested on ImageNet, and OOD is averaged over $5$ ImageNet variations. Experiments with $[1,2,4]$-shots are presented. Each accuracy is averaged over $3$ runs. We can see that our proposed method overwhelms WISE-FT in all cases.}
    \label{fig:id_ood_curve_avg_ft}
\end{figure}

Although learning with visual descriptors significantly outperforms linear probing in the few-shot setting, we should remark that ImageNet and its variations are usually considered ``in-distribution'' to the CLIP training data. In this case, the zero-shot model itself is usually a very strong baseline, and typically outperforms few-shot models, as can be observed by comparing the results in \cref{table:zs_comparison} and \cref{table:fs_comparison}. WISE-FT serves as a strong method to improve both in-distribution and out-of-distribution accuracies.  We can apply WISE-FT to any of our existing settings, including SLR-AVD and LP. In particular, we can train a linear head (and/or image encoder, depending on the setting) $\mW\_{learned}$, and interpolate with the zero-shot weight  $\mW\_{zs}$ by taking a convex combination $\alpha\mW\_{zs}+(1-\alpha)\mW\_{learned}$, for $\alpha\in\{\alpha_1,\ldots, \alpha_n\}$. We are free to vary $\alpha$. Then for each $\alpha_i$, we can plot that weight ensemble's ID and OOD test accuracy. This procedure thus creates an ID-OOD frontier and along the curve, some ensemble excels at both ID and OOD distribution. 
We further show that WISE-FT+SLR-SVD dominates WISE-FT+LP. 
% We further show that with WISE-FT, our method still dominates the WISE-FT with image features. \anna{Is this like a WISE-FT of SLR-AVD vs a WISE-FT of VD? Consistent terminology here might help.} 
See the ID-OOD curves in \cref{fig:id_ood_curve_avg_ft}. We show the plot of $k=4,8,16$. SLR-AVD's ID-OOD curve overwhelms that of LP, indicating that SLR-AVD is better at both ID and OOD tasks.

\subsection{Comparison to finetuning}
% \anna{What does this sentence mean? Specifically the "WISE-FT the whole image encoder" Also WISE-SLR isn't defined} 
We compare WISE-FT where we additionally interpolate the image encoder, to WISE-SLR, an interpolation between SLR-FT-AVD and ZS-AVD.
% We compare WISE-FT the whole image encoder to WISE-SLR. 
The ID-OOD frontier is presented in \cref{fig:id_ood_curve_avg_ft} and the accuracies are reported in \cref{table:optimal_alpha_ft}.  

On the ID task, WISE-SLR outperforms vanilla WISE-FT by $1.43\%$, $1.62\%$, and $1.61\%$ respectively with $k=1,2,4$ shot training data. Averaging over $5$ distribution shift datasets, with optimal $\alpha$, WISE-SLR outperforms vanilla WISE-FT by $0.88\%$, $0.73\%$, and $0.64\%$ respectively for $k=1,2,4$. The optimal $\alpha$ is picked independently for each method on each dataset.

\subsection{Comparison to CoOp}
We compare linear probing with AVD to CoOp \citep{zhou2022learning} as well. CoOp learns the prefix of ``\texttt{[prefix] \{classname\}}'' in the continuous token embedding space. The benefit of CoOp is that it operates in a continuous space, hence one can optimize using standard backpropagation, and it is quite computationally efficient. On the other hand, due to the requirement of backprop, CoOp stores a large computation graph, hence memory-efficiency is a big advantage of SLR-AVD over CoOp.

When implementing CoOp, we choose a prefix of length $16$ and do not use a suffix. The prefix is fixed for all classes. We train with Adam for 20 epochs, setting the batch size to 512. This gives us comparable results to those of the original paper.

\begin{wraptable}{r}{0.55\linewidth}
% \begin{minipage}{\linewidth}
% \begin{table}[tb]
    % \vspace{-1em}
    \centering
    \caption{Accuracies of CoOp and SLR-AVD on ImageNet. For SLR-AVD, we interpolate the learned weights with zero-shot weights by $\alpha \mW\_{avd} + (1-\alpha) \mW\_{zs}$, where $\alpha=0.05k$ and $k$ is shots per class.}
    \begin{tabular}{c|ccccc}
\toprule
		 Shots & 1 & 2 & 4 & 8 & 16\\
		\midrule
		CoOp & 59.54 & 60.23 & 60.97 & 62.20 & 63.02 \\
		AVD & 61.27 & 61.72 & 62.52 & 63.73 & 64.97  \\
		$\Delta$ & +1.73 & +1.49 & +1.55 & +1.53 & +1.95\\
		 \bottomrule
    \end{tabular}
    
    \label{table:avd_vs_coop}
    % \end{table}
% \end{minipage}
\end{wraptable}

Since CoOp injects ``\texttt{classname}'' to the prompt during inference directly, this enforces a very strong prior. For a fair comparison, we also inject a strong prior by interpolating our learned linear head $\alpha \mW\_{avd} + (1-\alpha) \mW\_{zs}$ with the zero-shot linear classifier. We use $\alpha=0.05k$ where $k$ is the number of shots. Resnet-50 vision backbone is used for both methods. See the result in \cref{table:avd_vs_coop}. SLR-AVD exceeds CoOp by $1.73$, $1.49$, $1.55$, $1.53$, $1.95$ on $k=1,2,4,8,16$, respectively.

%\begin{table}[h]
%\centering
% \caption{Accuracies of CoOp and SLR-AVD on ImageNet. For SLR-AVD, we interpolate the learned weights with zero-shot weights by $\alpha \mW\_{avd} + (1-\alpha) \mW\_{zs}$, where $\alpha=0.05k$ and $k$ is the number of shot per class.}
% \begin{tabular}{c|ccccc} 
% \midrule
% Shots & 1 & 2 & 4 & 8 & 16\\
%\midrule
%CoOp & 59.54 & 60.23 & 60.97 & 62.20 & 63.02 \\
%AVD & 61.27 & 61.72 & 62.52 & 63.73 & 64.97  \\
%$\Delta$ & +1.73 & +1.49 & +1.55 & +1.53 & +1.95\\
%
% \bottomrule
% \end{tabular}
%
% \label{table:avd_vs_coop}
%\end{table}

\section{Conclusion}
Motivated by the invariance principle and information bottleneck, we present how to leverage descriptive features for image learning in the few-shot setting robustly. These descriptive features can be easily obtained from LLMs. Applying sparse logistic regression then successfully selects the important features, which turn out to be intuitive. Our proposed method outperforms linear probing and standard finetuning in both ID and OOD tasks, with or without combining with WISE-FT. This approach helps us further understand the CLIP embedding space and how the semantics serve as a strong robust prior. Moving forward, it is important to understand and quantify the robustness of the visual descriptors' space and compare it to the image embedding space statistically. From the practical side, this work aligns image encoders to a fixed text encoder; it is valuable to study how to simultaneously align both encoders in a robust way. 

\bibliography{example_paper}

\begin{thebibliography}{44}
\providecommand{\natexlab}[1]{#1}
\providecommand{\url}[1]{\texttt{#1}}
\expandafter\ifx\csname urlstyle\endcsname\relax
  \providecommand{\doi}[1]{doi: #1}\else
  \providecommand{\doi}{doi: \begingroup \urlstyle{rm}\Url}\fi

\bibitem[Ahuja et~al.(2021)Ahuja, Caballero, Zhang, Gagnon-Audet, Bengio,
  Mitliagkas, and Rish]{ahuja2021invariance}
Kartik Ahuja, Ethan Caballero, Dinghuai Zhang, Jean-Christophe Gagnon-Audet,
  Yoshua Bengio, Ioannis Mitliagkas, and Irina Rish.
\newblock Invariance principle meets information bottleneck for
  out-of-distribution generalization.
\newblock \emph{Advances in Neural Information Processing Systems},
  34:\penalty0 3438--3450, 2021.

\bibitem[Arjovsky et~al.(2019)Arjovsky, Bottou, Gulrajani, and
  Lopez-Paz]{arjovsky2019invariant}
Martin Arjovsky, L{\'e}on Bottou, Ishaan Gulrajani, and David Lopez-Paz.
\newblock Invariant risk minimization.
\newblock \emph{arXiv preprint arXiv:1907.02893}, 2019.

\bibitem[Arora et~al.(2018)Arora, Ge, Neyshabur, and Zhang]{arora2018stronger}
Sanjeev Arora, Rong Ge, Behnam Neyshabur, and Yi~Zhang.
\newblock Stronger generalization bounds for deep nets via a compression
  approach.
\newblock In \emph{International Conference on Machine Learning}, pp.\
  254--263. PMLR, 2018.

\bibitem[Barbu et~al.(2019)Barbu, Mayo, Alverio, Luo, Wang, Gutfreund,
  Tenenbaum, and Katz]{barbu2019objectnet}
Andrei Barbu, David Mayo, Julian Alverio, William Luo, Christopher Wang, Dan
  Gutfreund, Josh Tenenbaum, and Boris Katz.
\newblock Objectnet: A large-scale bias-controlled dataset for pushing the
  limits of object recognition models.
\newblock \emph{Advances in neural information processing systems}, 32, 2019.

\bibitem[Blier \& Ollivier(2018)Blier and Ollivier]{blier2018description}
L{\'e}onard Blier and Yann Ollivier.
\newblock The description length of deep learning models.
\newblock \emph{Advances in Neural Information Processing Systems}, 31, 2018.

\bibitem[Blum \& Langford(2003)Blum and Langford]{blum2003pac}
Avrim Blum and John Langford.
\newblock Pac-mdl bounds.
\newblock In \emph{Learning Theory and Kernel Machines: 16th Annual Conference
  on Learning Theory and 7th Kernel Workshop, COLT/Kernel 2003, Washington, DC,
  USA, August 24-27, 2003. Proceedings}, pp.\  344--357. Springer, 2003.

\bibitem[Chen et~al.(2022)Chen, Yao, Song, Li, Rao, and Zhang]{chen2022prompt}
Guangyi Chen, Weiran Yao, Xiangchen Song, Xinyue Li, Yongming Rao, and Kun
  Zhang.
\newblock Prompt learning with optimal transport for vision-language models.
\newblock \emph{arXiv preprint arXiv:2210.01253}, 2022.

\bibitem[Cover(1999)]{cover1999elements}
Thomas~M Cover.
\newblock \emph{Elements of information theory}.
\newblock John Wiley \& Sons, 1999.

\bibitem[Defazio et~al.(2014)Defazio, Bach, and
  Lacoste-Julien]{defazio2014saga}
Aaron Defazio, Francis Bach, and Simon Lacoste-Julien.
\newblock Saga: A fast incremental gradient method with support for
  non-strongly convex composite objectives.
\newblock \emph{Advances in neural information processing systems}, 27, 2014.

\bibitem[Deng et~al.(2009)Deng, Dong, Socher, Li, Li, and
  Fei-Fei]{deng2009imagenet}
Jia Deng, Wei Dong, Richard Socher, Li-Jia Li, Kai Li, and Li~Fei-Fei.
\newblock Imagenet: A large-scale hierarchical image database.
\newblock In \emph{2009 IEEE conference on computer vision and pattern
  recognition}, pp.\  248--255. Ieee, 2009.

\bibitem[Feng et~al.(2021)Feng, Han, and Du]{feng2021provable}
Zhili Feng, Shaobo Han, and Simon~S Du.
\newblock Provable adaptation across multiway domains via representation
  learning.
\newblock \emph{arXiv preprint arXiv:2106.06657}, 2021.

\bibitem[Galloway et~al.(2022)Galloway, Golubeva, Salem, Nica, Ioannou, and
  Taylor]{galloway2022bounding}
Angus Galloway, Anna Golubeva, Mahmoud Salem, Mihai Nica, Yani Ioannou, and
  Graham~W Taylor.
\newblock Bounding generalization error with input compression: An empirical
  study with infinite-width networks.
\newblock \emph{arXiv preprint arXiv:2207.09408}, 2022.

\bibitem[Goyal et~al.(2022)Goyal, Kumar, Garg, Kolter, and
  Raghunathan]{goyal2022finetune}
Sachin Goyal, Ananya Kumar, Sankalp Garg, Zico Kolter, and Aditi Raghunathan.
\newblock Finetune like you pretrain: Improved finetuning of zero-shot vision
  models.
\newblock \emph{arXiv preprint arXiv:2212.00638}, 2022.

\bibitem[Hendrycks et~al.(2021{\natexlab{a}})Hendrycks, Basart, Mu, Kadavath,
  Wang, Dorundo, Desai, Zhu, Parajuli, Guo, et~al.]{hendrycks2021many}
Dan Hendrycks, Steven Basart, Norman Mu, Saurav Kadavath, Frank Wang, Evan
  Dorundo, Rahul Desai, Tyler Zhu, Samyak Parajuli, Mike Guo, et~al.
\newblock The many faces of robustness: A critical analysis of
  out-of-distribution generalization.
\newblock In \emph{Proceedings of the IEEE/CVF International Conference on
  Computer Vision}, pp.\  8340--8349, 2021{\natexlab{a}}.

\bibitem[Hendrycks et~al.(2021{\natexlab{b}})Hendrycks, Zhao, Basart,
  Steinhardt, and Song]{hendrycks2021natural}
Dan Hendrycks, Kevin Zhao, Steven Basart, Jacob Steinhardt, and Dawn Song.
\newblock Natural adversarial examples.
\newblock In \emph{Proceedings of the IEEE/CVF Conference on Computer Vision
  and Pattern Recognition}, pp.\  15262--15271, 2021{\natexlab{b}}.

\bibitem[Khashabi et~al.(2021)Khashabi, Lyu, Min, Qin, Richardson, Singh,
  Welleck, Hajishirzi, Khot, Sabharwal, et~al.]{khashabi2021prompt}
Daniel Khashabi, Shane Lyu, Sewon Min, Lianhui Qin, Kyle Richardson, Sameer
  Singh, Sean Welleck, Hannaneh Hajishirzi, Tushar Khot, Ashish Sabharwal,
  et~al.
\newblock Prompt waywardness: The curious case of discretized interpretation of
  continuous prompts.
\newblock \emph{arXiv preprint arXiv:2112.08348}, 2021.

\bibitem[Koh et~al.(2021)Koh, Sagawa, Marklund, Xie, Zhang, Balsubramani, Hu,
  Yasunaga, Phillips, Gao, et~al.]{koh2021wilds}
Pang~Wei Koh, Shiori Sagawa, Henrik Marklund, Sang~Michael Xie, Marvin Zhang,
  Akshay Balsubramani, Weihua Hu, Michihiro Yasunaga, Richard~Lanas Phillips,
  Irena Gao, et~al.
\newblock Wilds: A benchmark of in-the-wild distribution shifts.
\newblock In \emph{International Conference on Machine Learning}, pp.\
  5637--5664. PMLR, 2021.

\bibitem[Krizhevsky et~al.(2009)Krizhevsky, Hinton,
  et~al.]{krizhevsky2009learning}
Alex Krizhevsky, Geoffrey Hinton, et~al.
\newblock Learning multiple layers of features from tiny images.
\newblock 2009.

\bibitem[Kumar et~al.(2022)Kumar, Raghunathan, Jones, Ma, and
  Liang]{kumar2022fine}
Ananya Kumar, Aditi Raghunathan, Robbie Jones, Tengyu Ma, and Percy Liang.
\newblock Fine-tuning can distort pretrained features and underperform
  out-of-distribution.
\newblock \emph{arXiv preprint arXiv:2202.10054}, 2022.

\bibitem[Li et~al.(2021)Li, Wang, Zhang, Li, Keutzer, Darrell, and
  Zhao]{li2021learning}
Bo~Li, Yezhen Wang, Shanghang Zhang, Dongsheng Li, Kurt Keutzer, Trevor
  Darrell, and Han Zhao.
\newblock Learning invariant representations and risks for semi-supervised
  domain adaptation.
\newblock In \emph{Proceedings of the IEEE/CVF Conference on Computer Vision
  and Pattern Recognition}, pp.\  1104--1113, 2021.

\bibitem[Li et~al.(2022)Li, Shen, Wang, Zhu, Li, Keutzer, and
  Zhao]{li2022invariant}
Bo~Li, Yifei Shen, Yezhen Wang, Wenzhen Zhu, Dongsheng Li, Kurt Keutzer, and
  Han Zhao.
\newblock Invariant information bottleneck for domain generalization.
\newblock In \emph{Proceedings of the AAAI Conference on Artificial
  Intelligence}, volume~36, pp.\  7399--7407, 2022.

\bibitem[Lu et~al.(2020)Lu, Nott, Olson, Todeschini, Vahabi, Carmon, and
  Schmidt]{lu2020harder}
Shangyun Lu, Bradley Nott, Aaron Olson, Alberto Todeschini, Hossein Vahabi,
  Yair Carmon, and Ludwig Schmidt.
\newblock Harder or different? a closer look at distribution shift in dataset
  reproduction.
\newblock In \emph{ICML Workshop on Uncertainty and Robustness in Deep
  Learning}, volume~5, pp.\ ~15, 2020.

\bibitem[Lu et~al.(2022)Lu, Liu, Zhang, Liu, and Tian]{lu2022prompt}
Yuning Lu, Jianzhuang Liu, Yonggang Zhang, Yajing Liu, and Xinmei Tian.
\newblock Prompt distribution learning.
\newblock In \emph{Proceedings of the IEEE/CVF Conference on Computer Vision
  and Pattern Recognition}, pp.\  5206--5215, 2022.

\bibitem[Menon \& Vondrick(2022)Menon and Vondrick]{menon2022visual}
Sachit Menon and Carl Vondrick.
\newblock Visual classification via description from large language models.
\newblock \emph{arXiv preprint arXiv:2210.07183}, 2022.

\bibitem[Moran \& Yehudayoff(2016)Moran and Yehudayoff]{moran2016sample}
Shay Moran and Amir Yehudayoff.
\newblock Sample compression schemes for vc classes.
\newblock \emph{Journal of the ACM (JACM)}, 63\penalty0 (3):\penalty0 1--10,
  2016.

\bibitem[Pearl(1995)]{pearl1995causal}
Judea Pearl.
\newblock Causal diagrams for empirical research.
\newblock \emph{Biometrika}, 82\penalty0 (4):\penalty0 669--688, 1995.

\bibitem[Pratt et~al.(2022)Pratt, Liu, and Farhadi]{pratt2022does}
Sarah Pratt, Rosanne Liu, and Ali Farhadi.
\newblock What does a platypus look like? generating customized prompts for
  zero-shot image classification.
\newblock \emph{arXiv preprint arXiv:2209.03320}, 2022.

\bibitem[Radford et~al.(2021)Radford, Kim, Hallacy, Ramesh, Goh, Agarwal,
  Sastry, Askell, Mishkin, Clark, et~al.]{radford2021learning}
Alec Radford, Jong~Wook Kim, Chris Hallacy, Aditya Ramesh, Gabriel Goh,
  Sandhini Agarwal, Girish Sastry, Amanda Askell, Pamela Mishkin, Jack Clark,
  et~al.
\newblock Learning transferable visual models from natural language
  supervision.
\newblock In \emph{International Conference on Machine Learning}, pp.\
  8748--8763. PMLR, 2021.

\bibitem[Recht et~al.(2018)Recht, Roelofs, Schmidt, and
  Shankar]{recht2018cifar}
Benjamin Recht, Rebecca Roelofs, Ludwig Schmidt, and Vaishaal Shankar.
\newblock Do cifar-10 classifiers generalize to cifar-10?
\newblock \emph{arXiv preprint arXiv:1806.00451}, 2018.

\bibitem[Recht et~al.(2019)Recht, Roelofs, Schmidt, and
  Shankar]{recht2019imagenet}
Benjamin Recht, Rebecca Roelofs, Ludwig Schmidt, and Vaishaal Shankar.
\newblock Do imagenet classifiers generalize to imagenet?
\newblock In \emph{International conference on machine learning}, pp.\
  5389--5400. PMLR, 2019.

\bibitem[Rosenfeld et~al.(2020)Rosenfeld, Ravikumar, and
  Risteski]{rosenfeld2020risks}
Elan Rosenfeld, Pradeep Ravikumar, and Andrej Risteski.
\newblock The risks of invariant risk minimization.
\newblock \emph{arXiv preprint arXiv:2010.05761}, 2020.

\bibitem[Shwartz-Ziv et~al.(2018)Shwartz-Ziv, Painsky, and
  Tishby]{shwartz2018representation}
Ravid Shwartz-Ziv, Amichai Painsky, and Naftali Tishby.
\newblock Representation compression and generalization in deep neural
  networks, 2018.

\bibitem[Song \& Ermon(2019)Song and Ermon]{song2019understanding}
Jiaming Song and Stefano Ermon.
\newblock Understanding the limitations of variational mutual information
  estimators.
\newblock \emph{arXiv preprint arXiv:1910.06222}, 2019.

\bibitem[Tishby \& Zaslavsky(2015)Tishby and Zaslavsky]{tishby2015deep}
Naftali Tishby and Noga Zaslavsky.
\newblock Deep learning and the information bottleneck principle.
\newblock In \emph{2015 ieee information theory workshop (itw)}, pp.\  1--5.
  IEEE, 2015.

\bibitem[Wang et~al.(2019)Wang, Ge, Lipton, and Xing]{wang2019learning}
Haohan Wang, Songwei Ge, Zachary Lipton, and Eric~P Xing.
\newblock Learning robust global representations by penalizing local predictive
  power.
\newblock \emph{Advances in Neural Information Processing Systems}, 32, 2019.

\bibitem[Wen et~al.(2023)Wen, Jain, Kirchenbauer, Goldblum, Geiping, and
  Goldstein]{wen2023hard}
Yuxin Wen, Neel Jain, John Kirchenbauer, Micah Goldblum, Jonas Geiping, and Tom
  Goldstein.
\newblock Hard prompts made easy: Gradient-based discrete optimization for
  prompt tuning and discovery.
\newblock \emph{arXiv preprint arXiv:2302.03668}, 2023.

\bibitem[Wong et~al.(2021)Wong, Santurkar, and Madry]{wong2021leveraging}
Eric Wong, Shibani Santurkar, and Aleksander Madry.
\newblock Leveraging sparse linear layers for debuggable deep networks.
\newblock In \emph{International Conference on Machine Learning}, pp.\
  11205--11216. PMLR, 2021.

\bibitem[Wortsman et~al.(2022)Wortsman, Ilharco, Kim, Li, Kornblith, Roelofs,
  Lopes, Hajishirzi, Farhadi, Namkoong, et~al.]{wortsman2022robust}
Mitchell Wortsman, Gabriel Ilharco, Jong~Wook Kim, Mike Li, Simon Kornblith,
  Rebecca Roelofs, Raphael~Gontijo Lopes, Hannaneh Hajishirzi, Ali Farhadi,
  Hongseok Namkoong, et~al.
\newblock Robust fine-tuning of zero-shot models.
\newblock In \emph{Proceedings of the IEEE/CVF Conference on Computer Vision
  and Pattern Recognition}, pp.\  7959--7971, 2022.

\bibitem[Yang et~al.(2022)Yang, Panagopoulou, Zhou, Jin, Callison-Burch, and
  Yatskar]{yang2022language}
Yue Yang, Artemis Panagopoulou, Shenghao Zhou, Daniel Jin, Chris
  Callison-Burch, and Mark Yatskar.
\newblock Language in a bottle: Language model guided concept bottlenecks for
  interpretable image classification.
\newblock \emph{arXiv preprint arXiv:2211.11158}, 2022.

\bibitem[Zhao et~al.(2019)Zhao, Des~Combes, Zhang, and
  Gordon]{zhao2019learning}
Han Zhao, Remi~Tachet Des~Combes, Kun Zhang, and Geoffrey Gordon.
\newblock On learning invariant representations for domain adaptation.
\newblock In \emph{International conference on machine learning}, pp.\
  7523--7532. PMLR, 2019.

\bibitem[Zhao et~al.(2022)Zhao, Dan, Aragam, Jaakkola, Gordon, and
  Ravikumar]{zhao2022fundamental}
Han Zhao, Chen Dan, Bryon Aragam, Tommi~S Jaakkola, Geoffrey~J Gordon, and
  Pradeep Ravikumar.
\newblock Fundamental limits and tradeoffs in invariant representation
  learning.
\newblock \emph{The Journal of Machine Learning Research}, 23\penalty0
  (1):\penalty0 15356--15404, 2022.

\bibitem[Zhou et~al.(2022{\natexlab{a}})Zhou, Yang, Loy, and
  Liu]{zhou2022conditional}
Kaiyang Zhou, Jingkang Yang, Chen~Change Loy, and Ziwei Liu.
\newblock Conditional prompt learning for vision-language models.
\newblock In \emph{Proceedings of the IEEE/CVF Conference on Computer Vision
  and Pattern Recognition}, pp.\  16816--16825, 2022{\natexlab{a}}.

\bibitem[Zhou et~al.(2022{\natexlab{b}})Zhou, Yang, Loy, and
  Liu]{zhou2022learning}
Kaiyang Zhou, Jingkang Yang, Chen~Change Loy, and Ziwei Liu.
\newblock Learning to prompt for vision-language models.
\newblock \emph{International Journal of Computer Vision}, 130\penalty0
  (9):\penalty0 2337--2348, 2022{\natexlab{b}}.

\bibitem[Zhu et~al.(2022)Zhu, Niu, Han, Wu, and Zhang]{zhu2022prompt}
Beier Zhu, Yulei Niu, Yucheng Han, Yue Wu, and Hanwang Zhang.
\newblock Prompt-aligned gradient for prompt tuning.
\newblock \emph{arXiv preprint arXiv:2205.14865}, 2022.

\end{thebibliography}
\bibliographystyle{iclr2024_conference}

\newpage
\appendix
\section{Appendix}

\paragraph{Hyperparameter} For ImageNet and its variations, we fix a set of 6804 augmented visual descriptors. The hyperparameters are swept over disjoint training and validation sets of size $20$ per class for LP and SLR-AVD. For $\ell_1$ regularization, its non-smoothness makes it notoriously hard for auto-differentiation. To circumvent the smoothness issue, we apply the GPU implementation \citep{wong2021leveraging} of a variance-reduction proximal gradient method SAGA \citep{defazio2014saga}.
We adopt the \textit{regularization path} approach, in which the solver optimizes over $100$ regularization strengths $\lambda_1>\lambda_2\cdots >\lambda_{100}$. Here we set $\lambda_1$ to be the strength that returns a model that uses none of the features, and $\lambda_{100}=0.1\times\lambda_1$. For LP, we always use $\ell_2$ regularization, we use L-BFGS implemented by scikit-learn, and search for the regularization strength over $100$ grids between $0.5$ and $6$. All the $\lambda$s are evenly spread in the log-space\footnote{In python numpy.logspace(math.log10($\lambda_1$), math.log10($\lambda_{100}$), 100)}. For FT and SLR-FT-AVD, we select hyperparameters using a training and validation set of size $4$ per class. The batch size is fixed to be $512$ and the number of epochs is fixed to be $10$. We always optimize with AdamW, and choose a cosine rate scheduler with warm-ups. We randomly select learning rate in $[1e-8, 3e-5]$, weight decay in $[0.1, 0.12]$, and warm up steps in $\{0, 50, 500\}$, for $20$ trials. The chosen parameters are then fixed throughout all experiments. 

\begin{table}[tb]
\centering
%\resizebox{\columnwidth}{!}{
 \begin{tabular}{c|l} 
 \midrule
\quad Classes & Features\\
\midrule
 \multirow{3}{*}{airplanes}& airplanes which has anticollision lights\\
& a photo of airplanes\\
& airplanes which has overhead storage bins\\ 
\midrule
 \multirow{3}{*}{cars}& cars which has body kit\\
& cars which has bumpers\\
& cars which has wheel arch trim\\ 
\midrule
 \multirow{3}{*}{birds}&  birds which has leg color\\
& birds which has flight silhouette\\
& birds which has eye color\\
\midrule
 \multirow{3}{*}{cats}& cats which has pink tongue\\
& cats which has pink nose\\
& cats which has slit pupils\\
\midrule
 \multirow{3}{*}{deer}& deer which has large facial glands\\
& deer which has long, tufted hair on the neck and shoulders\\
& deer which has short, curved antlers\\
\midrule
 \multirow{3}{*}{dogs}& dogs which has silky fur\\
& dogs which has pattern\\
& dogs which has floppy ears\\
\midrule
 \multirow{3}{*}{frogs}& frogs which has large, bulging eyes\\
& frogs which has ridged or wartylooking skin\\
& frogs which has a fold of skin along the back\\
\midrule
 \multirow{3}{*}{horses}& horses which has hooves\\
& horses which has temperament\\
& horses which has intelligence\\
\midrule
 \multirow{3}{*}{ships}&  ships which has lifeboats\\
& ships which has bridge\\
& ships which has bow\\
\midrule
 \multirow{3}{*}{trucks}& trucks which has trailersway control\\
& trucks which has grille\\
& trucks which has lift kits\\
\bottomrule
 \end{tabular}
% }
 \caption{Features selected when trained with $\ell_{1}$ norm on CIFAR-10. The selected important features for each class are intuitive. Notice that the feature selection method does not restrict the candidates to be that particular class's descriptors.}
 \label{table:cifar_features}
\end{table}

As a recap, we use the following acronyms for different methods and datasets:

\paragraph{ZS:} Zero-shot classification using text embeddings of hand-crafted prompts ensembles. 

\paragraph{ZS-VD, ZS-AVD:} Zero-shot classification using visual descriptor and augmented visual descriptors, respectively.

\paragraph{LP:} Linear probing using image embeddings.

\paragraph{SLR-AVD:} Sparse logistic regression using AVDs. 

\paragraph{FT:} Finetuning the image encoder and classification head.

\paragraph{SLR-FT-AVD:} Sparse logistic regression with AVD, and then finetune the linear head plus the image encoder with frozen sparsity patterns.

\paragraph{WISE-FT:} Weight ensemble using ZS and FT.

\paragraph{WISE-SLR:} Weight ensemble using SLR-FT-AVD and ZS-AVD. 

\paragraph{IN:} ImageNet.

\paragraph{IN-R:} ImageNet-R.

\paragraph{IN-A:} ImageNet-A.

\paragraph{IN-V2:} ImageNetV2.

\paragraph{IN-Sketch:} ImageNet-Sketch.

\begin{figure}
     \centering
     \begin{subfigure}[b]{0.3\textwidth}
         \centering
         \includegraphics[width=\textwidth]{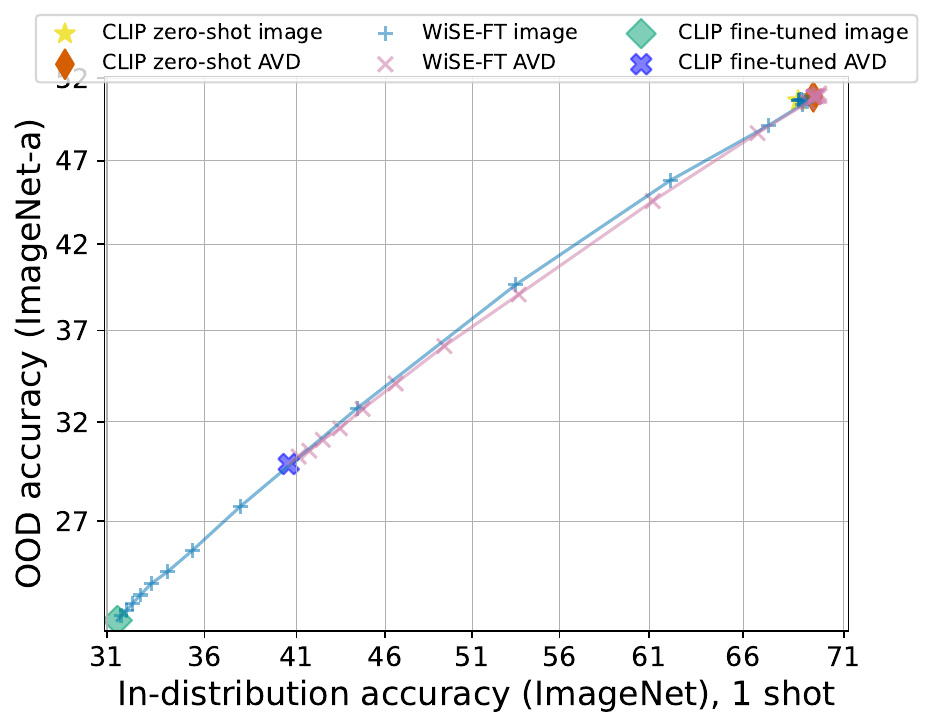}
     \end{subfigure}
     \begin{subfigure}[b]{0.3\textwidth}
         \centering
         \includegraphics[width=\textwidth]{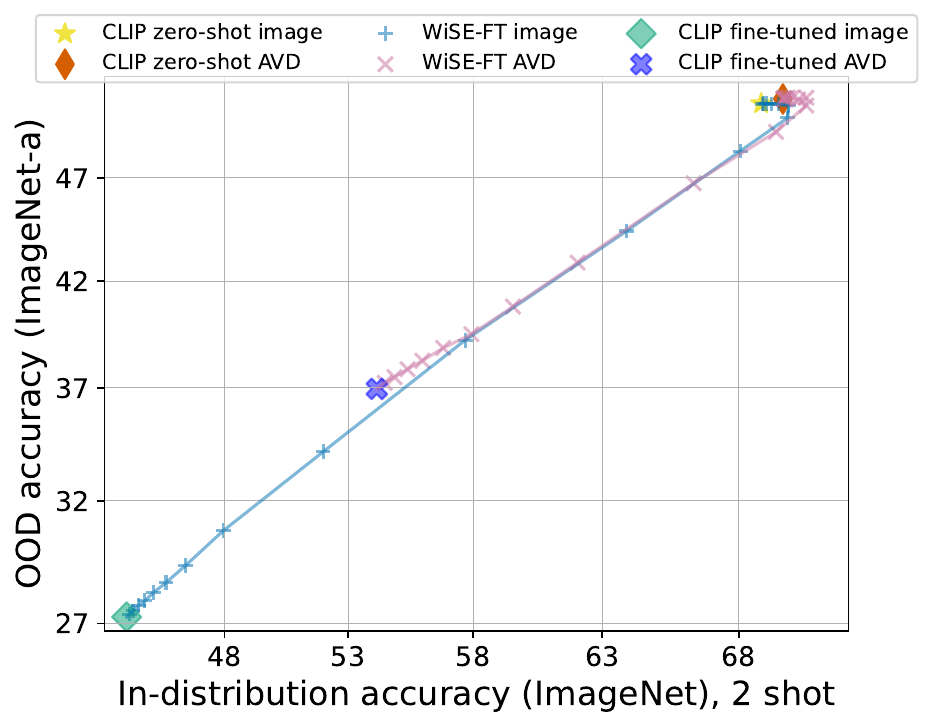}
     \end{subfigure}
     \begin{subfigure}[b]{0.3\textwidth}
         \centering
         \includegraphics[width=\textwidth]{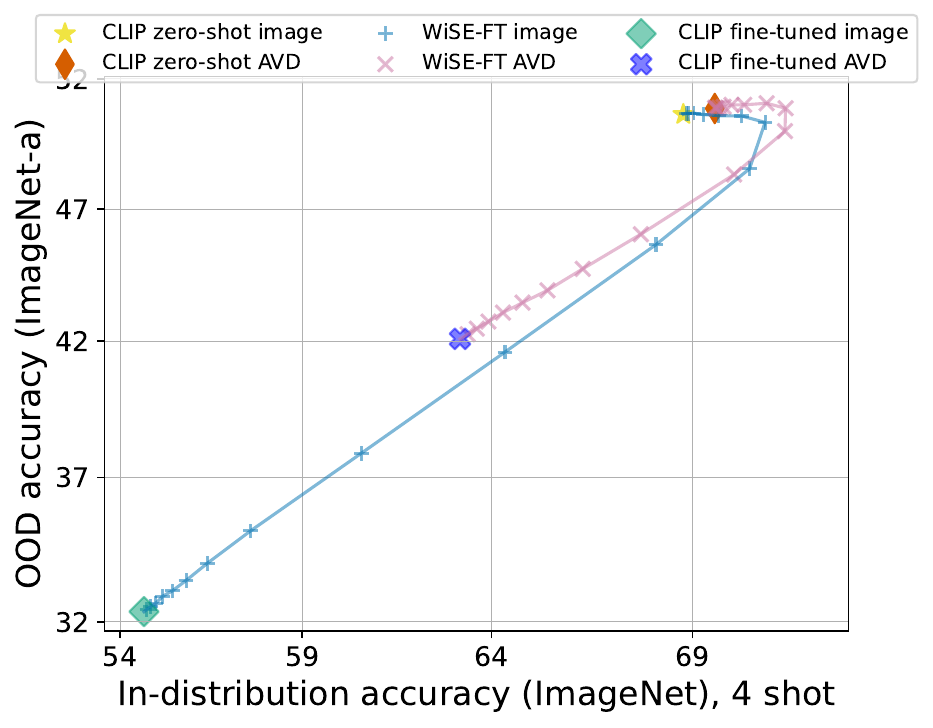}
     \end{subfigure}
     \hfill
     
     \begin{subfigure}[b]{0.3\textwidth}
         \centering
         \includegraphics[width=\textwidth]{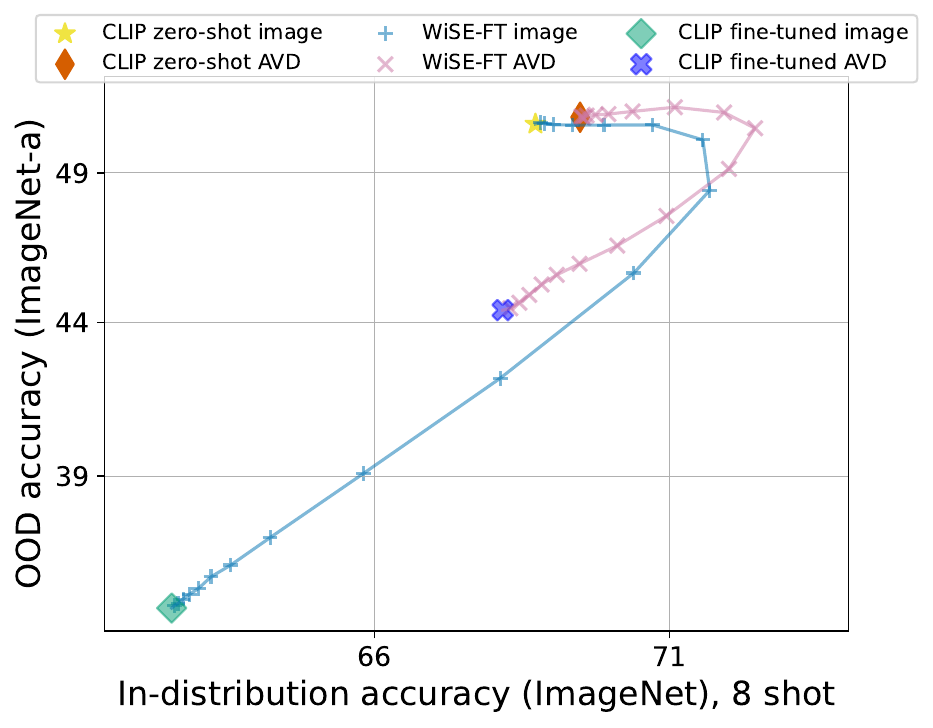}
     \end{subfigure}
     \begin{subfigure}[b]{0.3\textwidth}
         \centering
         \includegraphics[width=\textwidth]{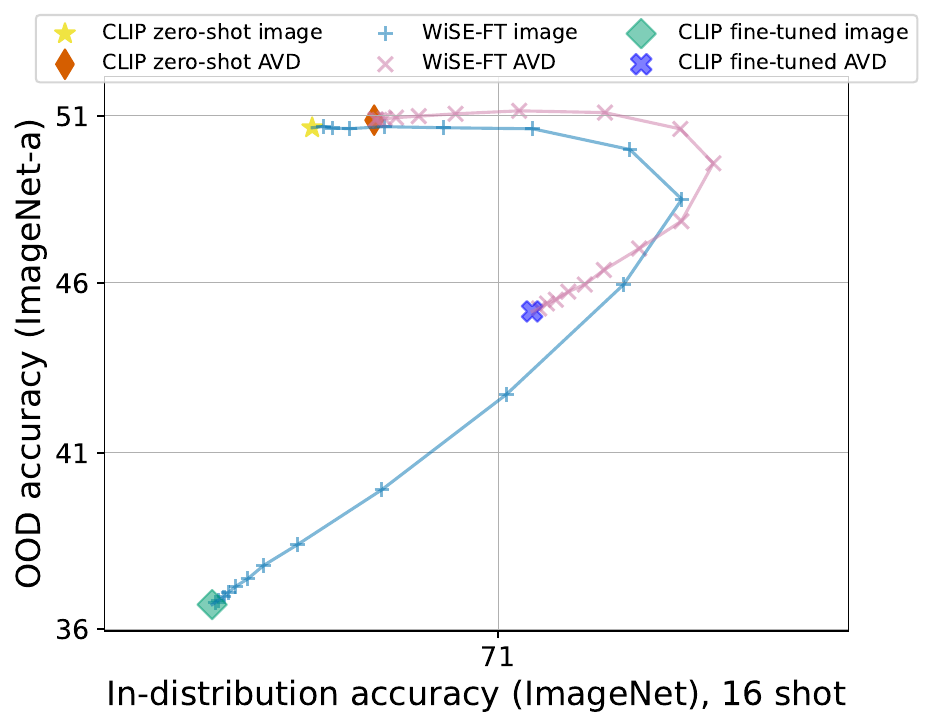}
     \end{subfigure}
     \begin{subfigure}[b]{0.3\textwidth}
         \centering
         \includegraphics[width=\textwidth]{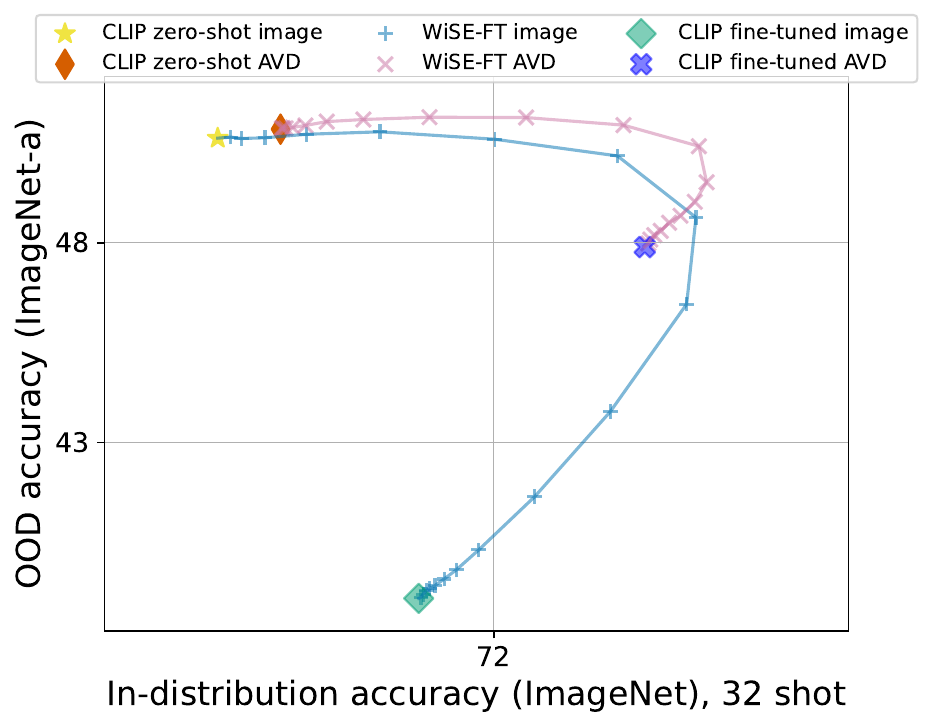}
     \end{subfigure}
    \caption{ID-OOD curves of LP vs SLR-AVD on IN-A. $k=1,2,4,8,16,32$.}
        \label{fig:id_ood_curve_in_a}
\end{figure}

\begin{figure}
     \centering
     \begin{subfigure}[b]{0.3\textwidth}
         \centering
         \includegraphics[width=\textwidth]{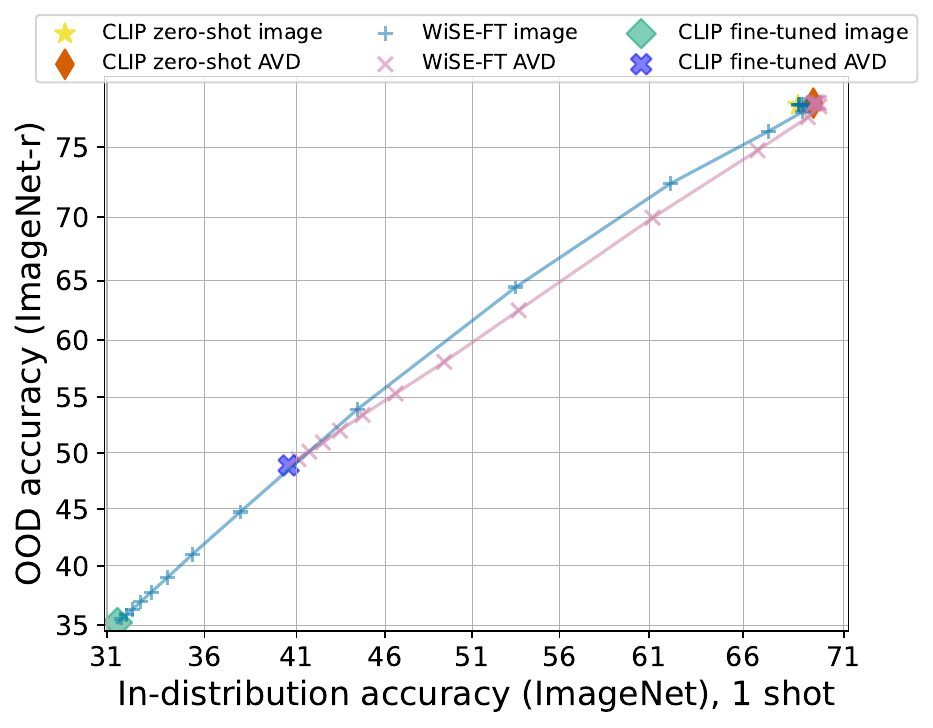}
     \end{subfigure}
     \begin{subfigure}[b]{0.3\textwidth}
         \centering
         \includegraphics[width=\textwidth]{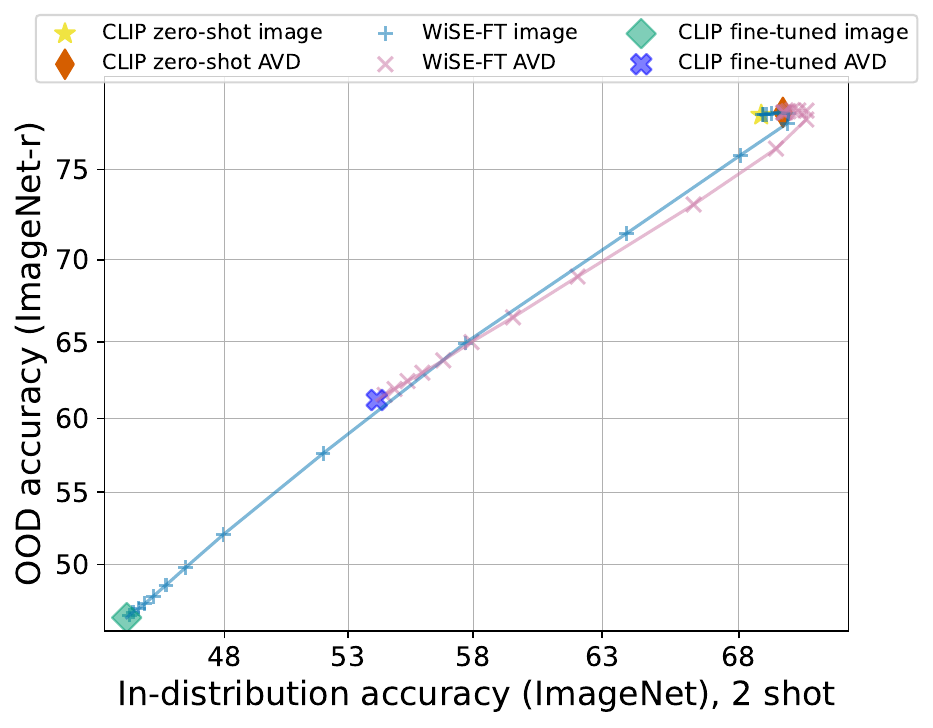}
     \end{subfigure}
     \begin{subfigure}[b]{0.3\textwidth}
         \centering
         \includegraphics[width=\textwidth]{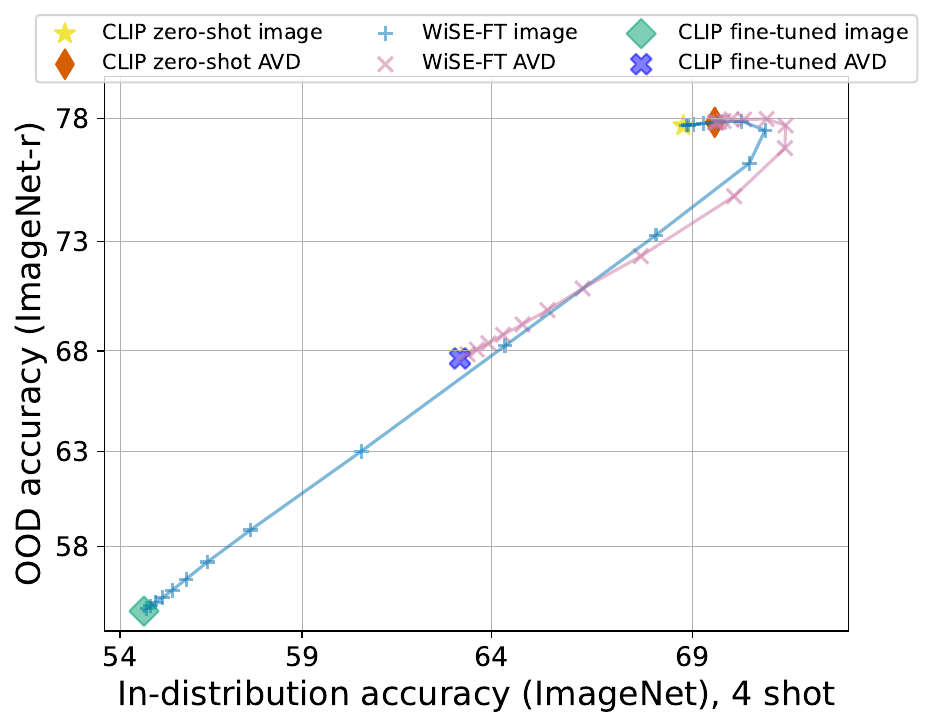}
     \end{subfigure}
     \hfill
     
     \begin{subfigure}[b]{0.3\textwidth}
         \centering
         \includegraphics[width=\textwidth]{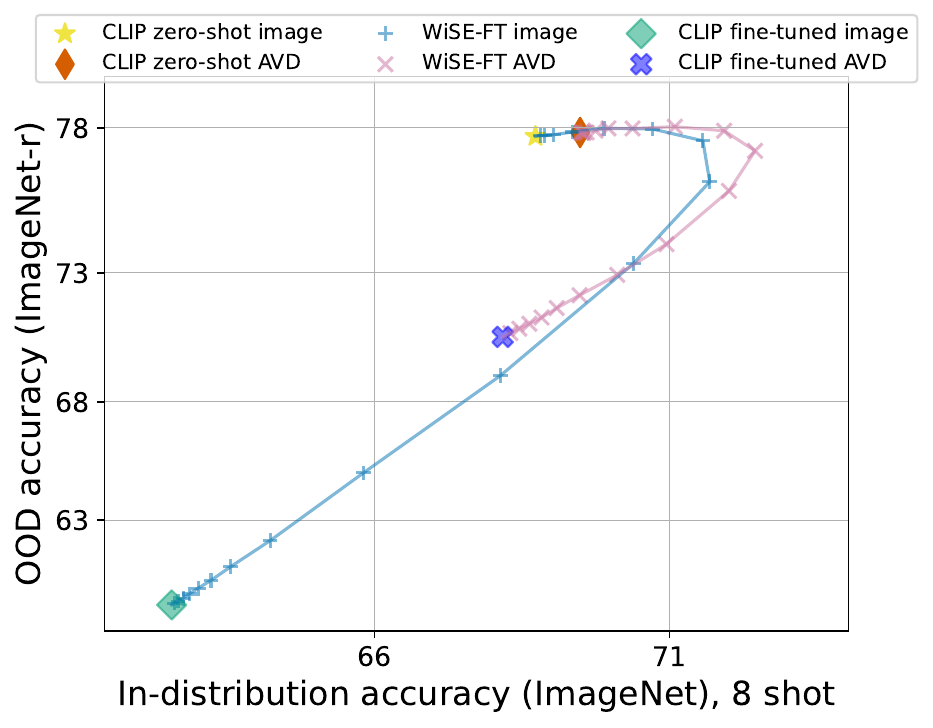}
     \end{subfigure}
     \begin{subfigure}[b]{0.3\textwidth}
         \centering
         \includegraphics[width=\textwidth]{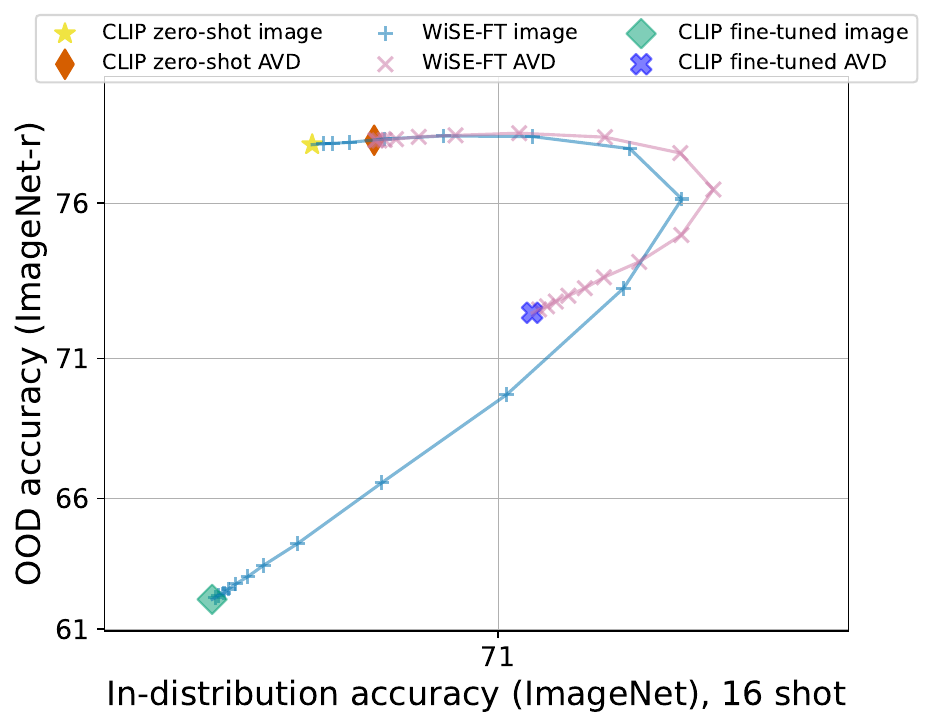}
     \end{subfigure}
     \begin{subfigure}[b]{0.3\textwidth}
         \centering
         \includegraphics[width=\textwidth]{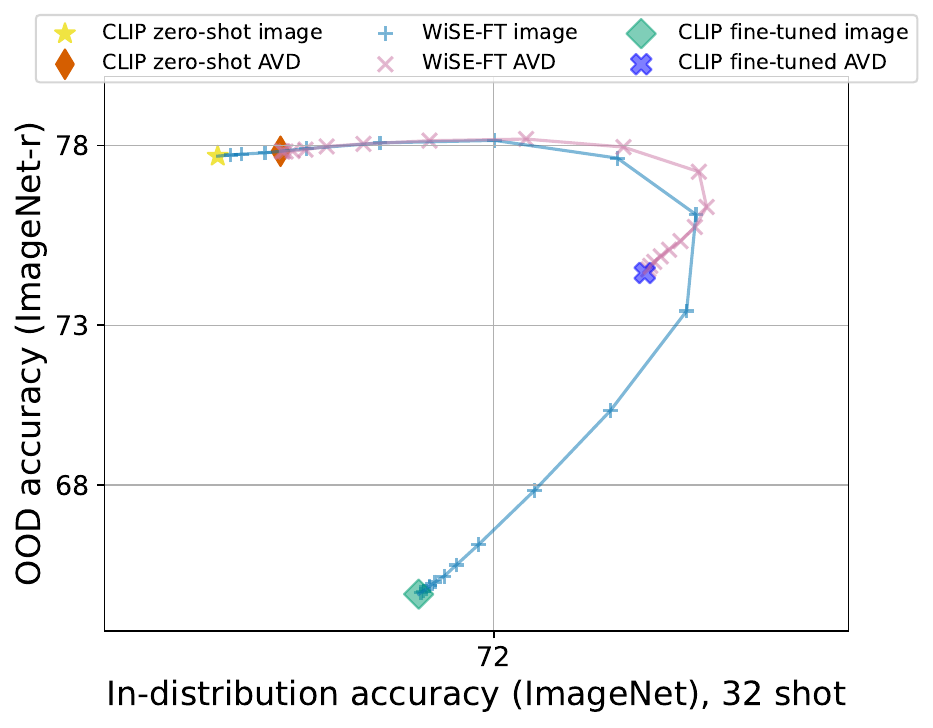}
     \end{subfigure}
    \caption{ID-OOD curves of LP vs SLR-AVD on IN-R. $k=1,2,4,8,16,32$.}
        \label{fig:id_ood_curve_in_r}
\end{figure}

\begin{figure}
     \centering
     \begin{subfigure}[b]{0.3\textwidth}
         \centering
         \includegraphics[width=\textwidth]{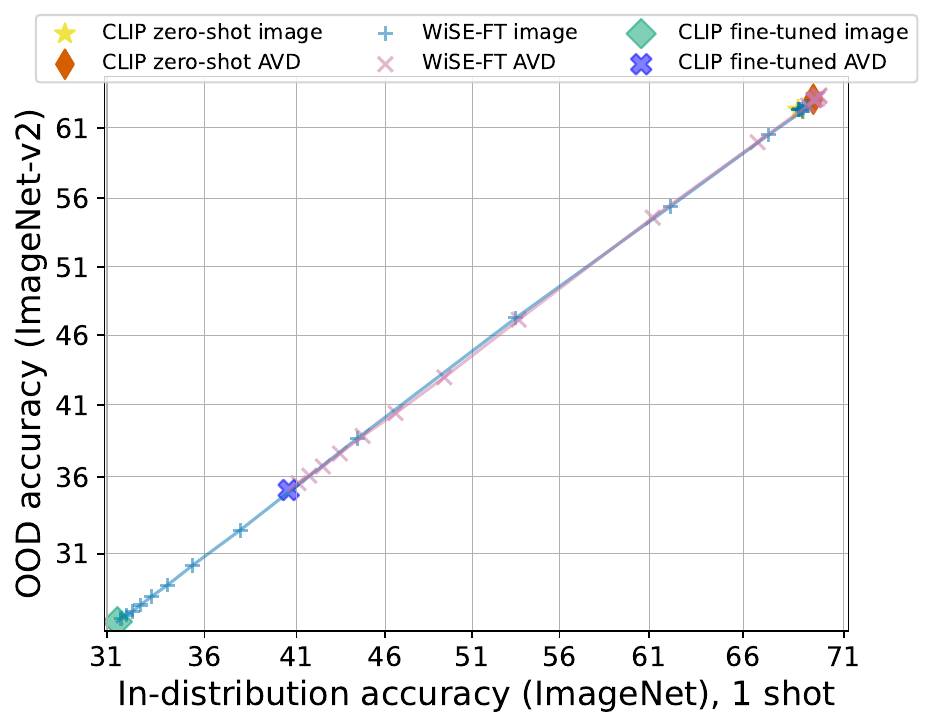}
     \end{subfigure}
     \begin{subfigure}[b]{0.3\textwidth}
         \centering
         \includegraphics[width=\textwidth]{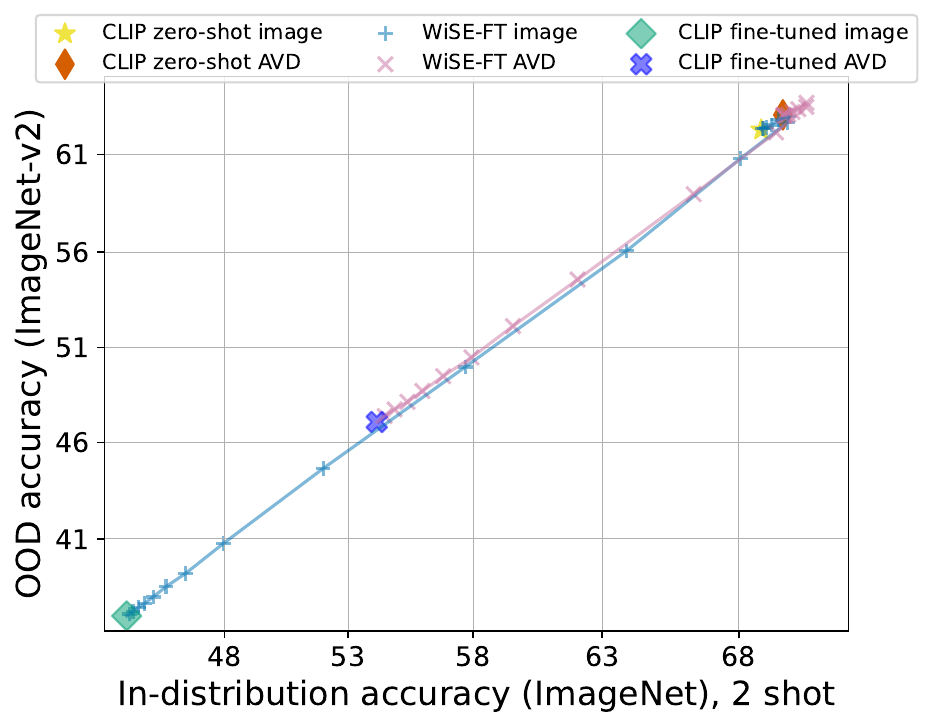}
     \end{subfigure}
     \begin{subfigure}[b]{0.3\textwidth}
         \centering
         \includegraphics[width=\textwidth]{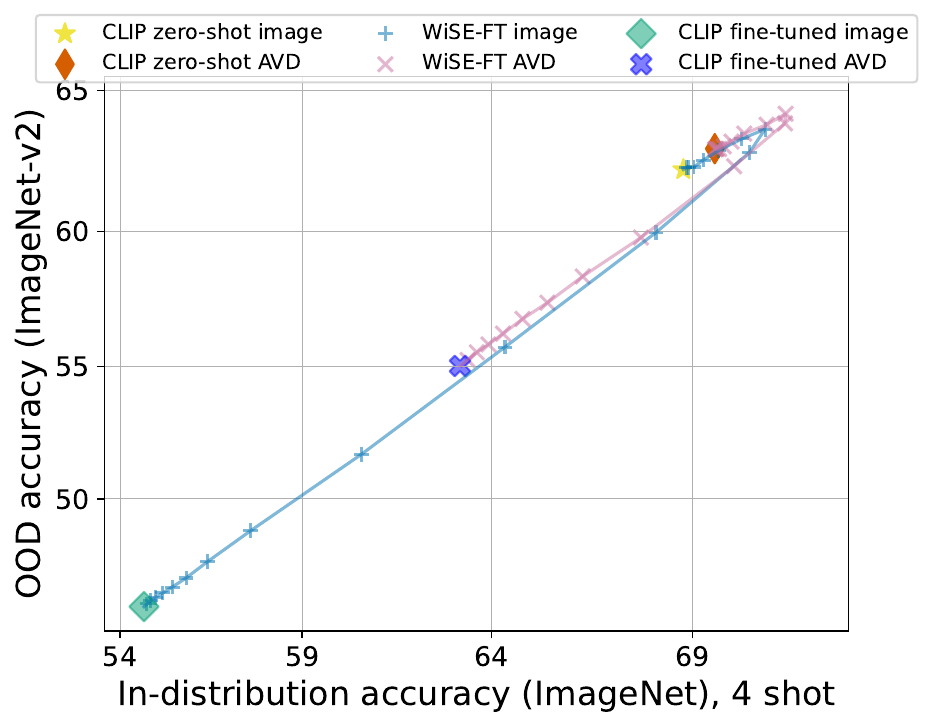}
     \end{subfigure}
     \hfill
     
     \begin{subfigure}[b]{0.3\textwidth}
         \centering
         \includegraphics[width=\textwidth]{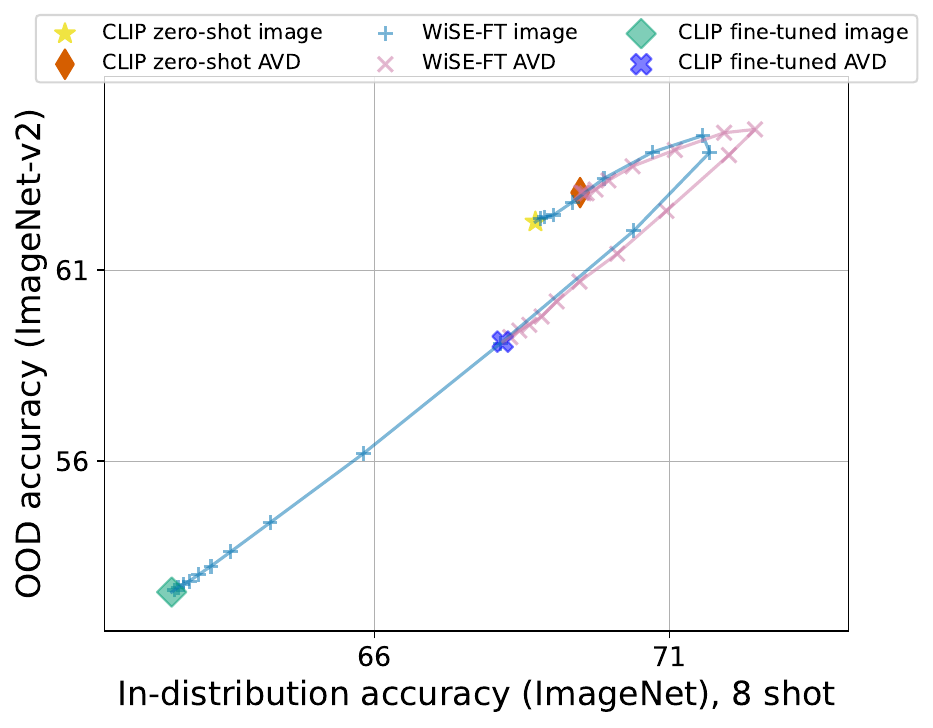}
     \end{subfigure}
     \begin{subfigure}[b]{0.3\textwidth}
         \centering
         \includegraphics[width=\textwidth]{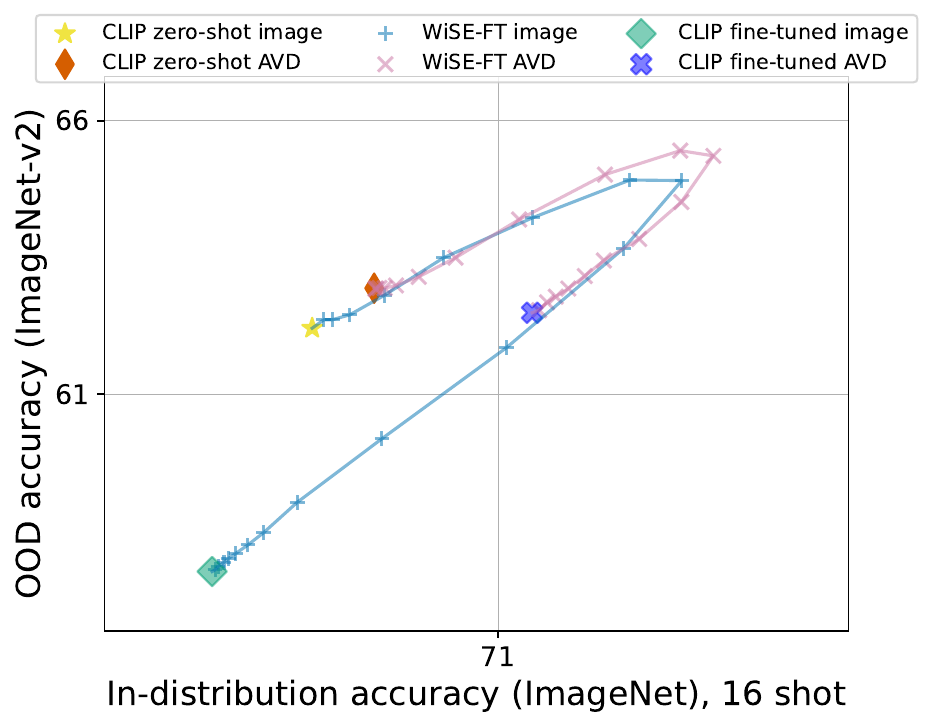}
     \end{subfigure}
     \begin{subfigure}[b]{0.3\textwidth}
         \centering
         \includegraphics[width=\textwidth]{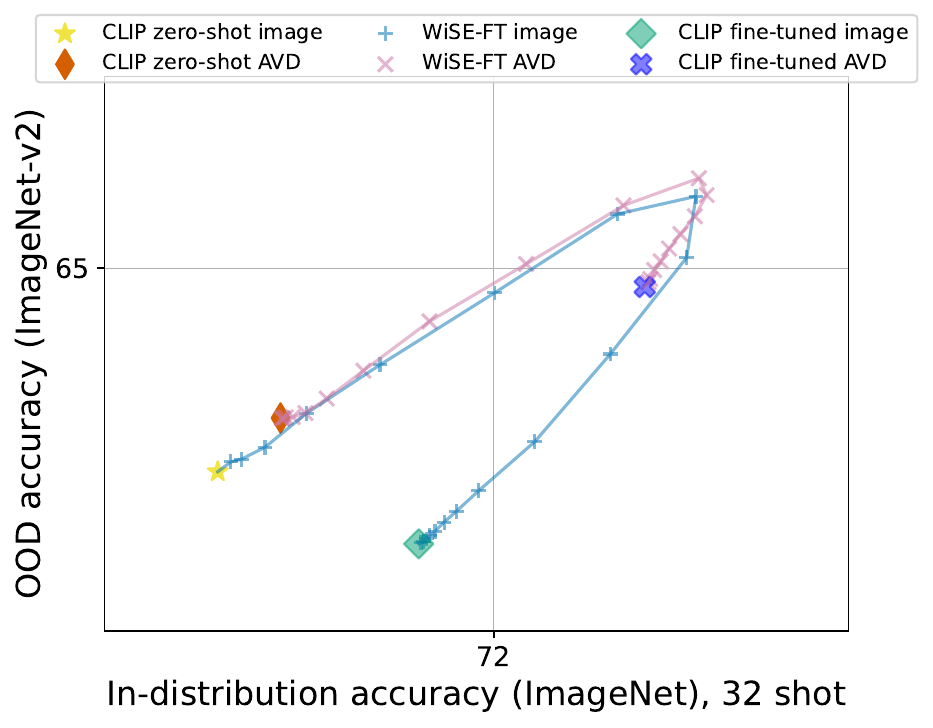}
     \end{subfigure}
    \caption{ID-OOD curves of LP vs SLR-AVD on IN-V2. $k=1,2,4,8,16,32$.}
        \label{fig:id_ood_curve_in_v2}
\end{figure}

\begin{figure}
     \centering
     \begin{subfigure}[b]{0.3\textwidth}
         \centering
         \includegraphics[width=\textwidth]{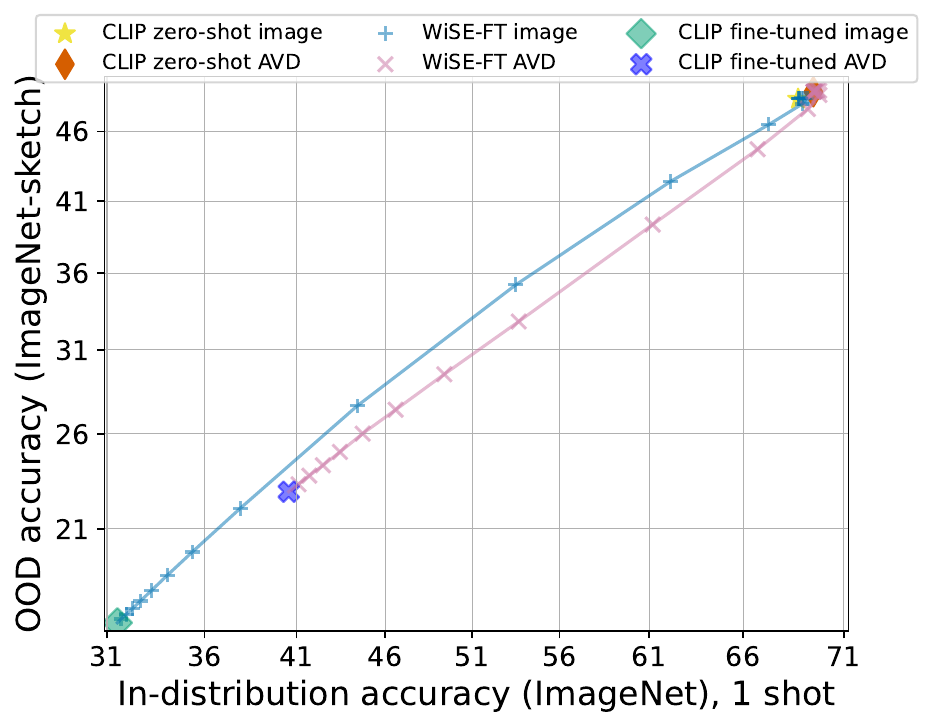}
     \end{subfigure}
     \begin{subfigure}[b]{0.3\textwidth}
         \centering
         \includegraphics[width=\textwidth]{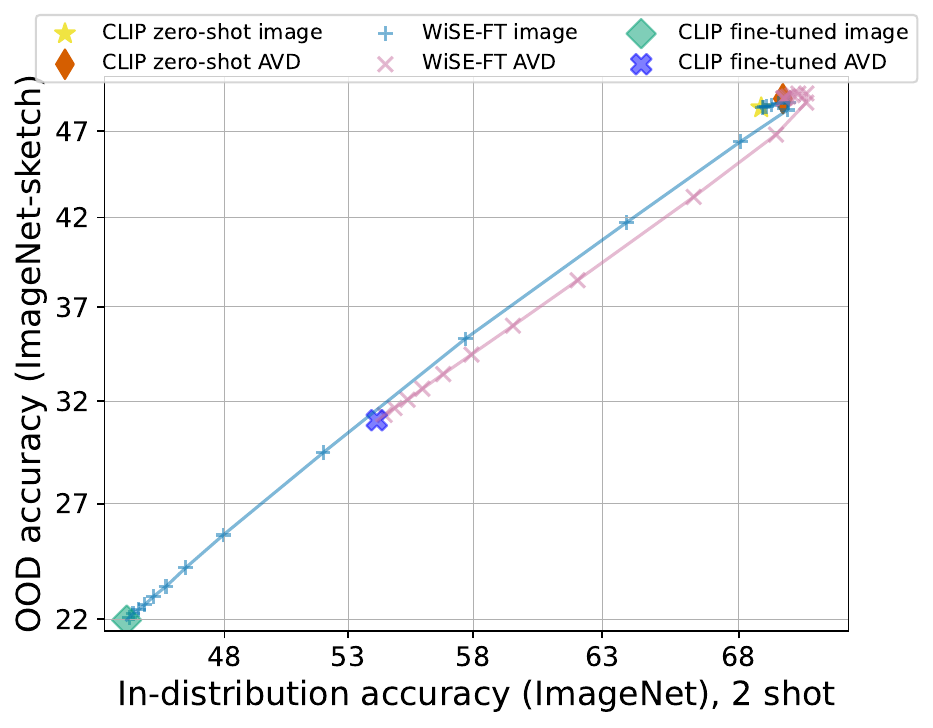}
     \end{subfigure}
     \begin{subfigure}[b]{0.3\textwidth}
         \centering
         \includegraphics[width=\textwidth]{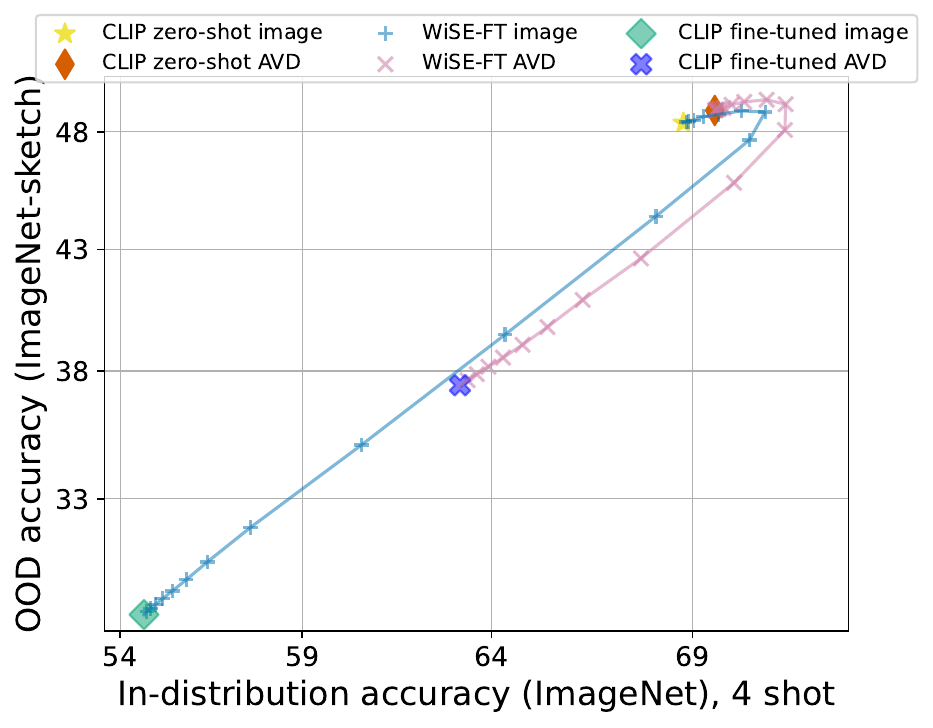}
     \end{subfigure}
     \hfill
     
     \begin{subfigure}[b]{0.3\textwidth}
         \centering
         \includegraphics[width=\textwidth]{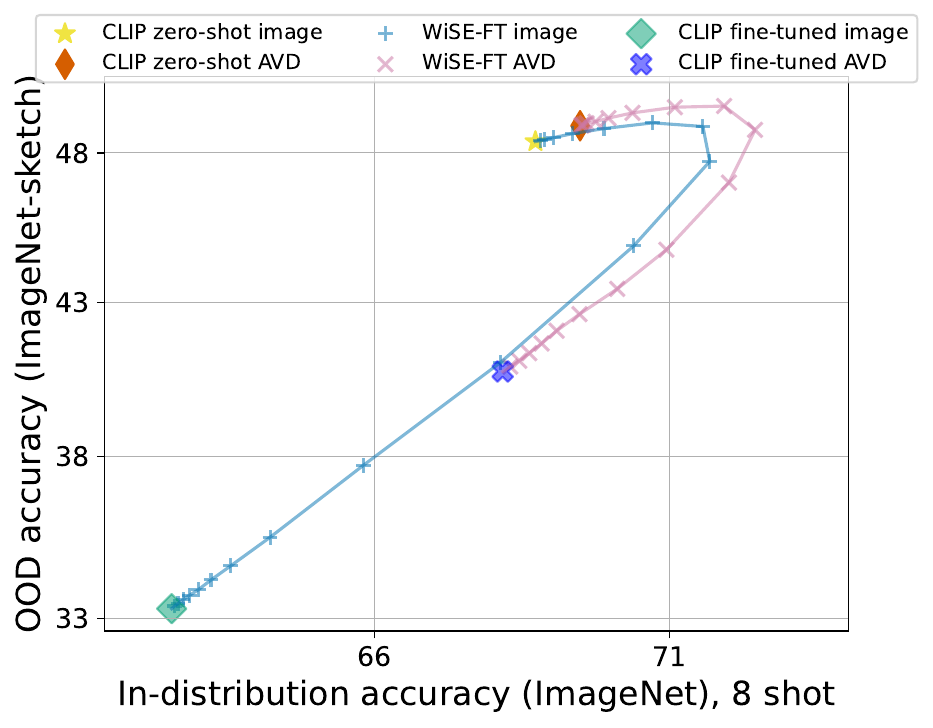}
     \end{subfigure}
     \begin{subfigure}[b]{0.3\textwidth}
         \centering
         \includegraphics[width=\textwidth]{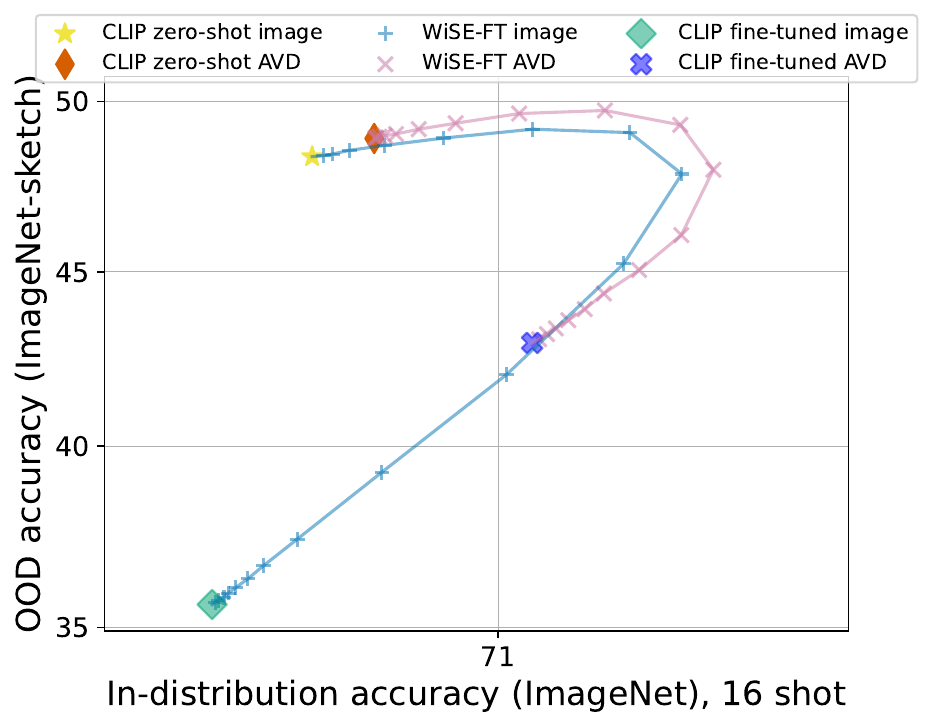}
     \end{subfigure}
     \begin{subfigure}[b]{0.3\textwidth}
         \centering
         \includegraphics[width=\textwidth]{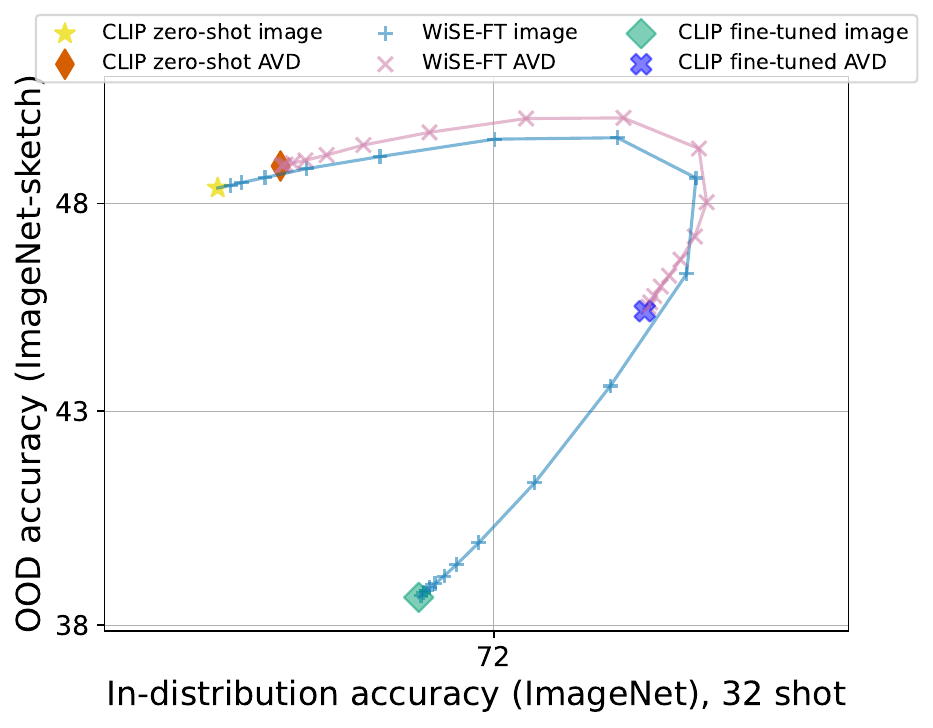}
     \end{subfigure}
    \caption{ID-OOD curves of LP vs SLR-AVD on IN-Sketch. $k=1,2,4,8,16,32$.}
        \label{fig:id_ood_curve_in_sketch}
\end{figure}

\begin{figure}
     \centering
     \begin{subfigure}[b]{0.3\textwidth}
         \centering
         \includegraphics[width=\textwidth]{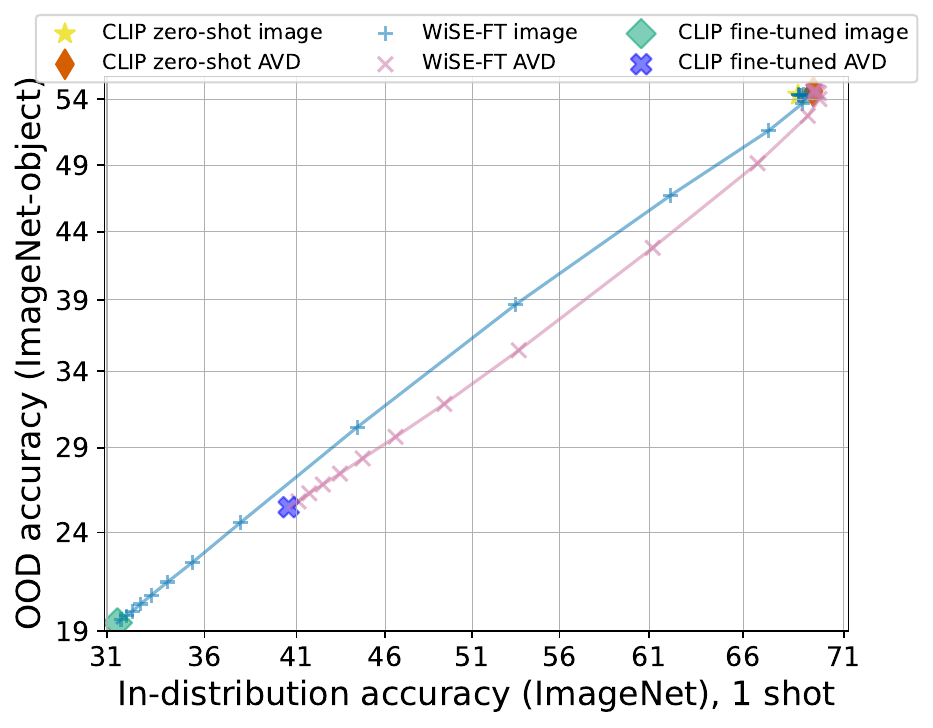}
     \end{subfigure}
     \begin{subfigure}[b]{0.3\textwidth}
         \centering
         \includegraphics[width=\textwidth]{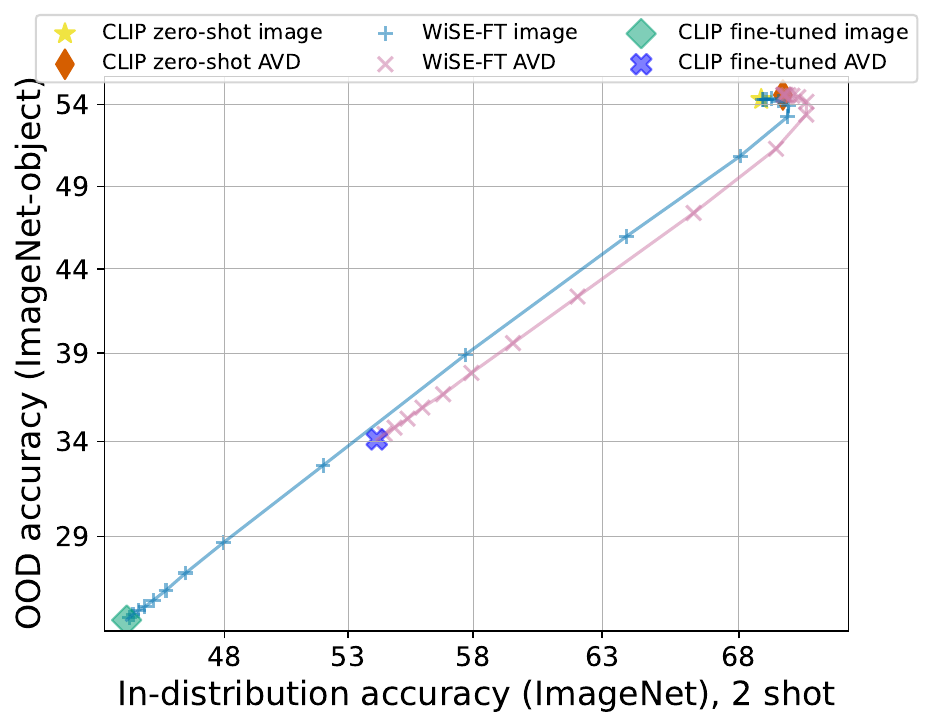}
     \end{subfigure}
     \begin{subfigure}[b]{0.3\textwidth}
         \centering
         \includegraphics[width=\textwidth]{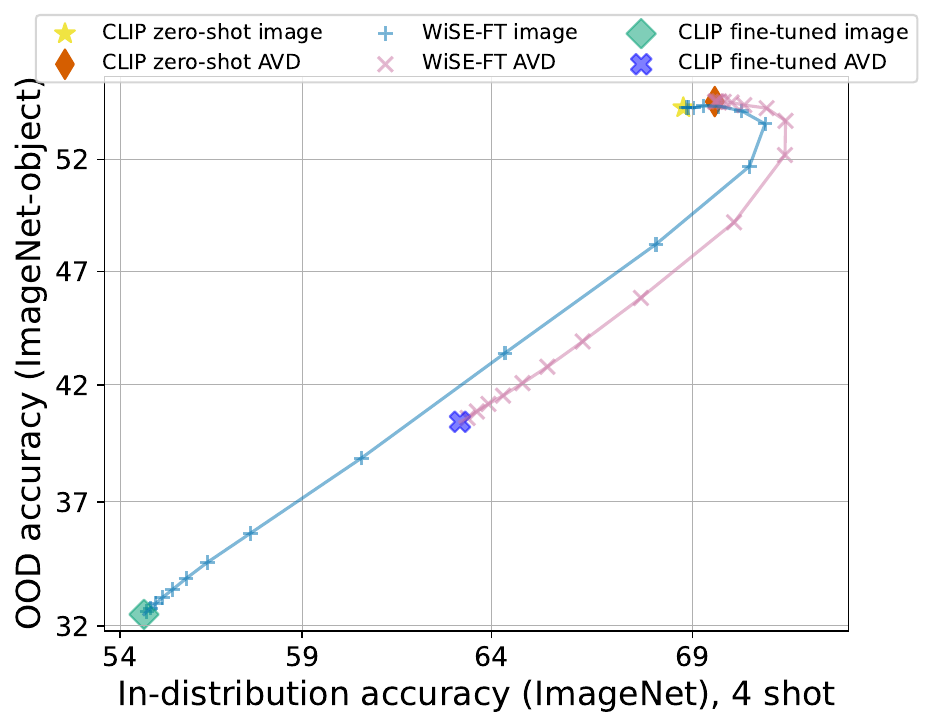}
     \end{subfigure}
     \hfill
     
     \begin{subfigure}[b]{0.3\textwidth}
         \centering
         \includegraphics[width=\textwidth]{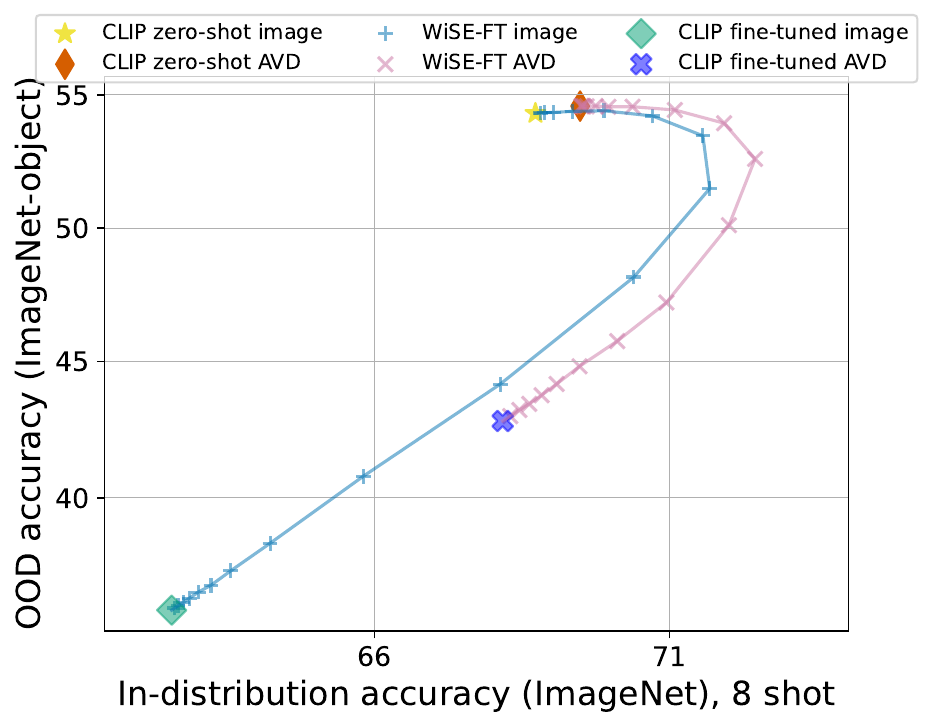}
     \end{subfigure}
     \begin{subfigure}[b]{0.3\textwidth}
         \centering
         \includegraphics[width=\textwidth]{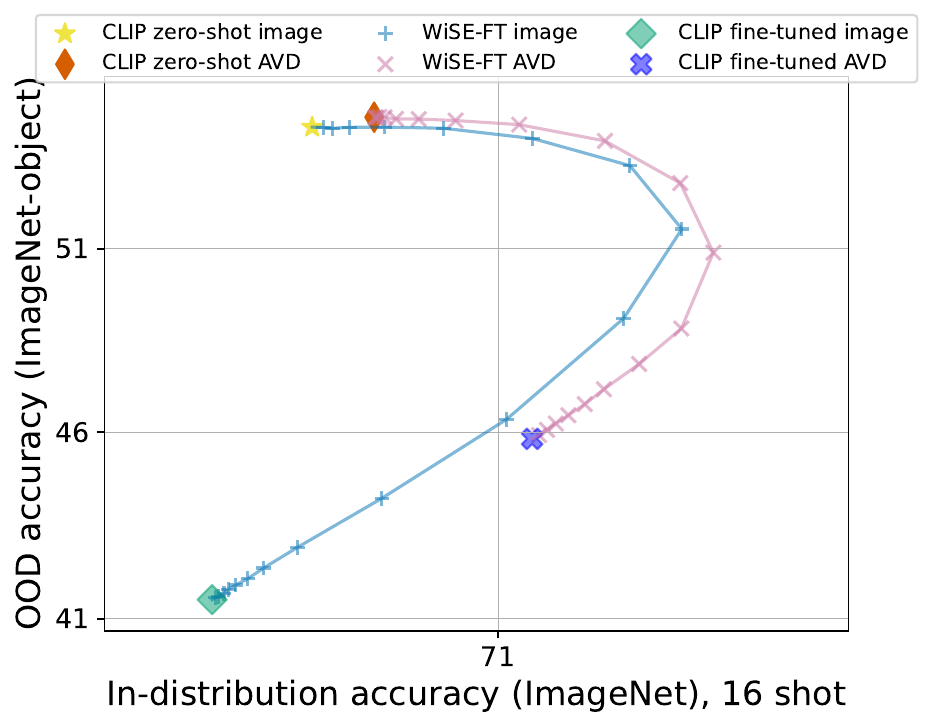}
     \end{subfigure}
     \begin{subfigure}[b]{0.3\textwidth}
         \centering
         \includegraphics[width=\textwidth]{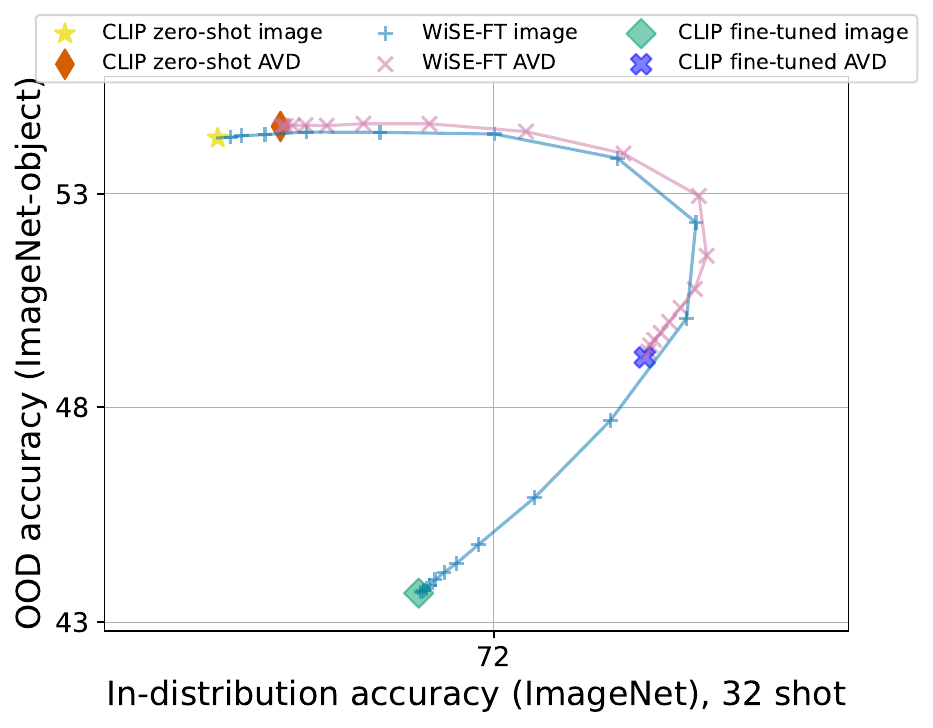}
     \end{subfigure}
    \caption{ID-OOD curves of LP vs SLR-AVD on ObjectNet. $k=1,2,4,8,16,32$.}
        \label{fig:id_ood_curve_in_object}
\end{figure}

The dataset-wise ID-OOD curves of LP vs SLR-AVD on IN-A, IN-R, IN-V2, IN-Sketch, and ObjectNet are listed in \cref{fig:id_ood_curve_in_a,fig:id_ood_curve_in_r,fig:id_ood_curve_in_v2,fig:id_ood_curve_in_sketch,fig:id_ood_curve_in_object}, respectively.

%%%%%%%%%%%%%%%%%%%. IMAGENET-A %%%%%%%%%%%%%%%%%%%%%

%%%%%%%%%%%%%%%%%%%. IMAGENET-R %%%%%%%%%%%%%%%%%%%%%

%%%%%%%%%%%%%%%%%%%. IMAGENET-V2 %%%%%%%%%%%%%%%%%%%%%

%%%%%%%%%%%%%%%%%%%. IMAGENET-SKETCH %%%%%%%%%%%%%%%%%%%%%

%%%%%%%%%%%%%%%%%%%. OBJECTNET %%%%%%%%%%%%%%%%%%%%%

\begin{figure}
     \centering
     \begin{subfigure}[b]{0.3\textwidth}
         \centering
         \includegraphics[width=\textwidth]{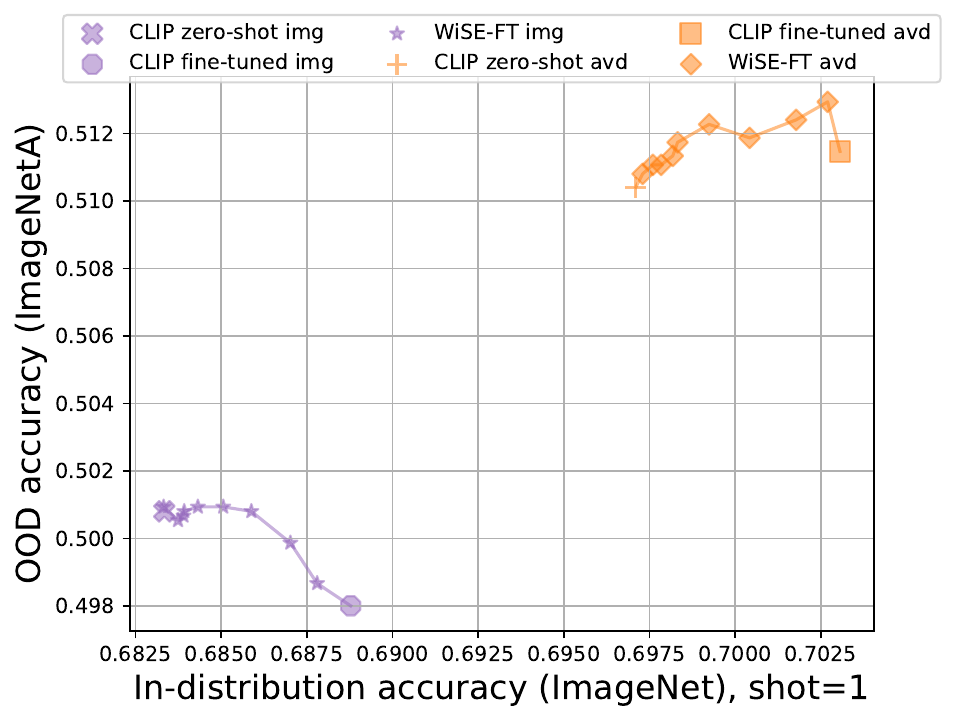}
     \end{subfigure}
     \begin{subfigure}[b]{0.3\textwidth}
         \centering
         \includegraphics[width=\textwidth]{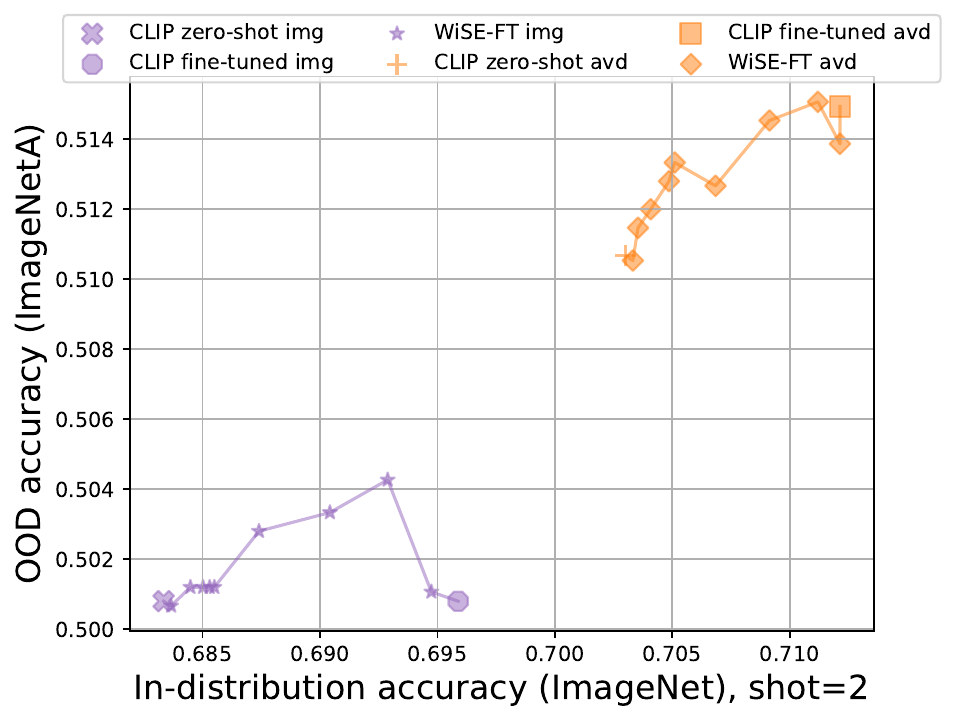}
     \end{subfigure}
     \begin{subfigure}[b]{0.3\textwidth}
         \centering
         \includegraphics[width=\textwidth]{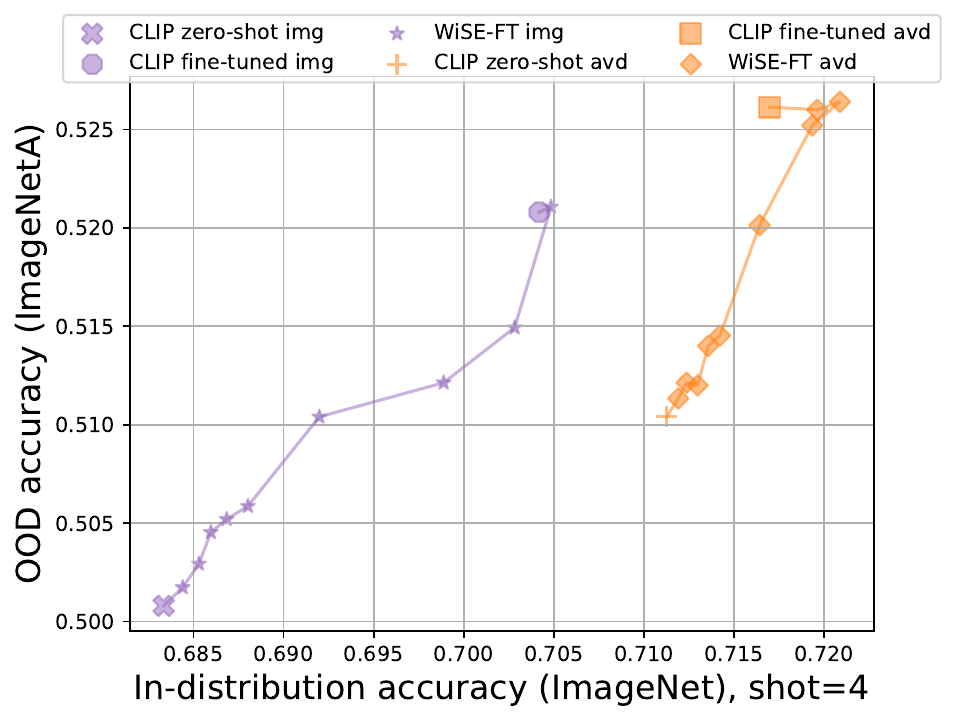}
     \end{subfigure}
    \caption{ID-OOD curves of WISE-FT vs WISE-SLR on IN-A. $k=1,2,4$.}
        \label{fig:id_ood_curve_in_a_ft}
\end{figure}

\begin{figure}
     \centering
     \begin{subfigure}[b]{0.3\textwidth}
         \centering
         \includegraphics[width=\textwidth]{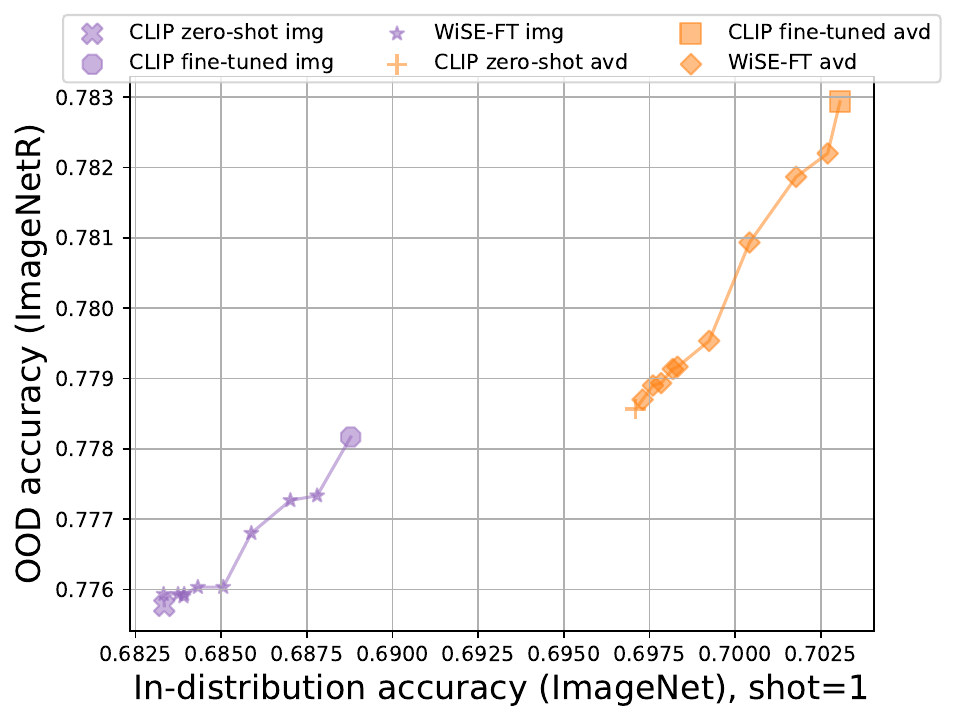}
     \end{subfigure}
     \begin{subfigure}[b]{0.3\textwidth}
         \centering
         \includegraphics[width=\textwidth]{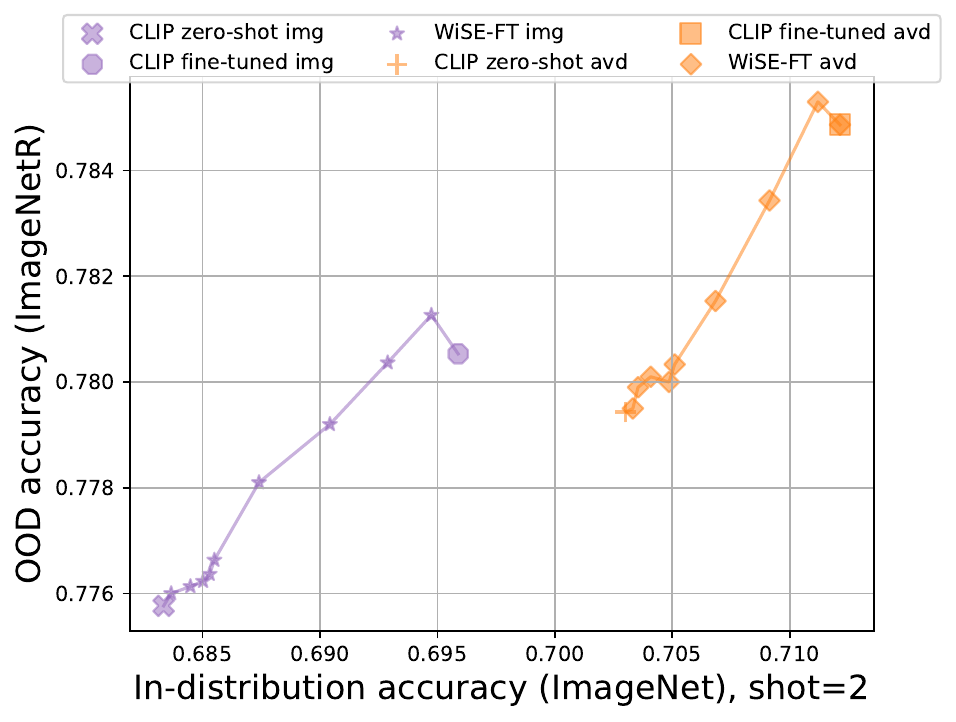}
     \end{subfigure}
     \begin{subfigure}[b]{0.3\textwidth}
         \centering
         \includegraphics[width=\textwidth]{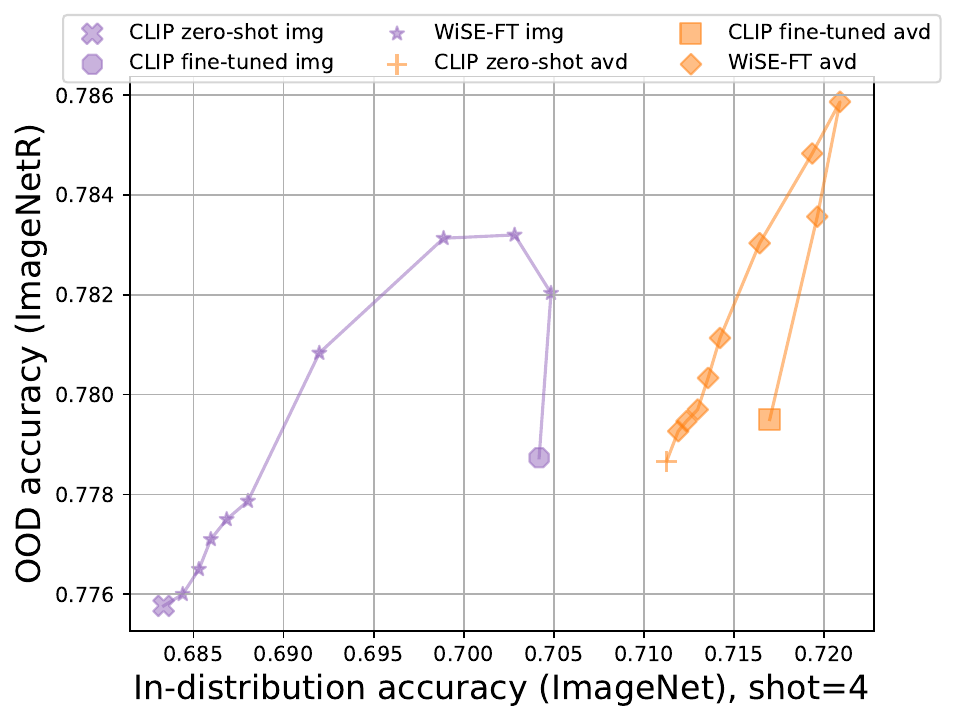}
     \end{subfigure}
    \caption{ID-OOD curves of WISE-FT vs WISE-SLR on IN-R. $k=1,2,4$.}
        \label{fig:id_ood_curve_in_r_ft}
\end{figure}

\begin{figure}
     \centering
     \begin{subfigure}[b]{0.3\textwidth}
         \centering
         \includegraphics[width=\textwidth]{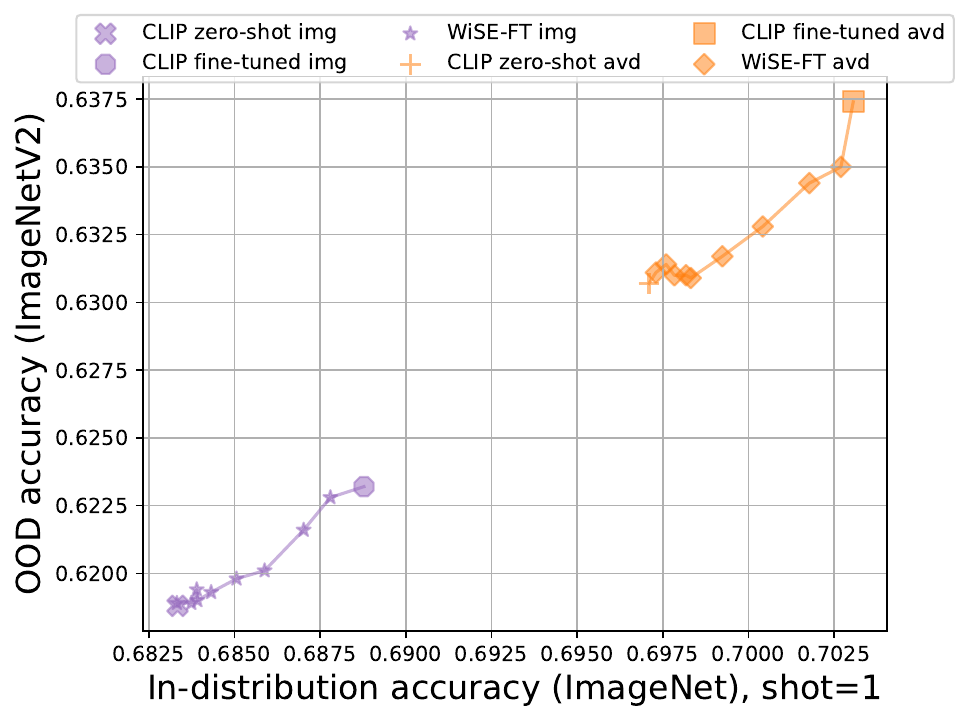}
     \end{subfigure}
     \begin{subfigure}[b]{0.3\textwidth}
         \centering
         \includegraphics[width=\textwidth]{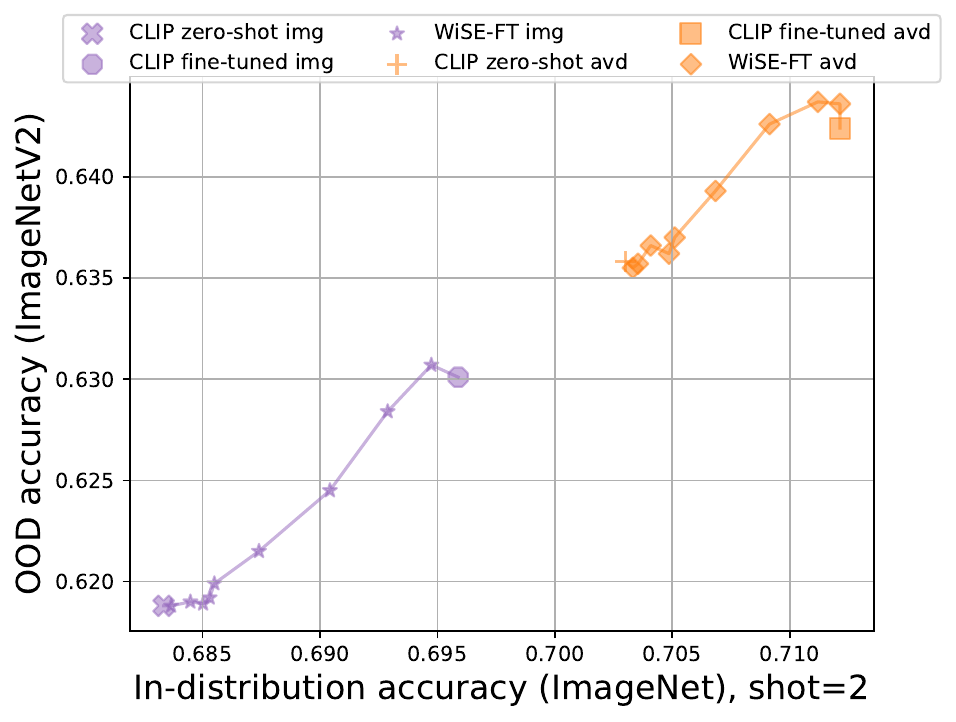}
     \end{subfigure}
     \begin{subfigure}[b]{0.3\textwidth}
         \centering
         \includegraphics[width=\textwidth]{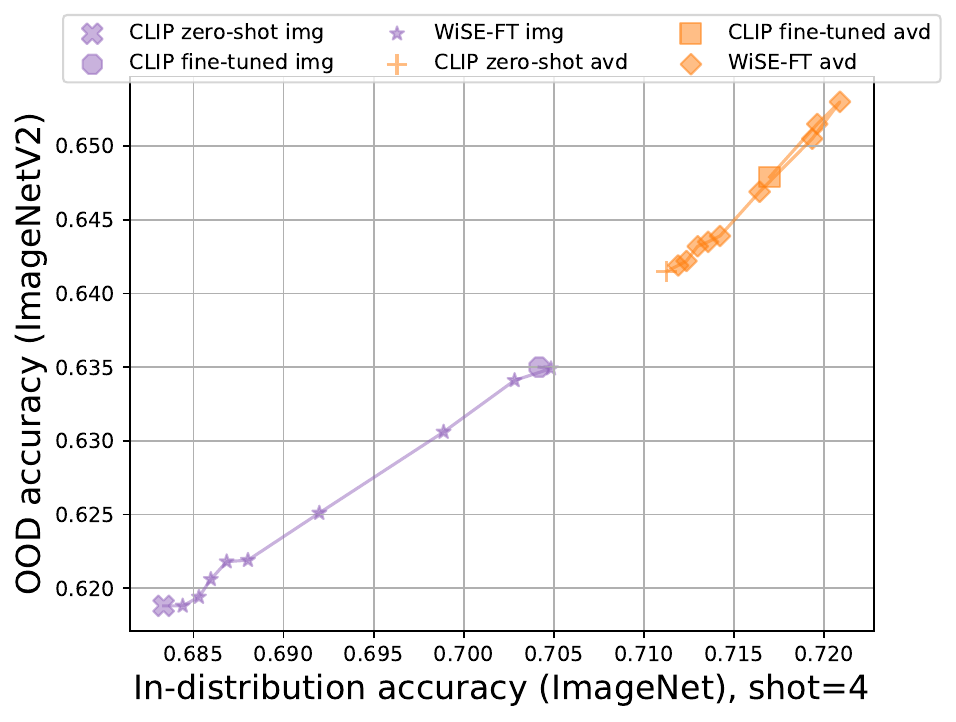}
     \end{subfigure}
    \caption{ID-OOD curves of WISE-FT vs WISE-SLR on IN-V2. $k=1,2,4$.}
        \label{fig:id_ood_curve_in_v2_ft}
\end{figure}

\begin{figure}
     \centering
     \begin{subfigure}[b]{0.3\textwidth}
         \centering
         \includegraphics[width=\textwidth]{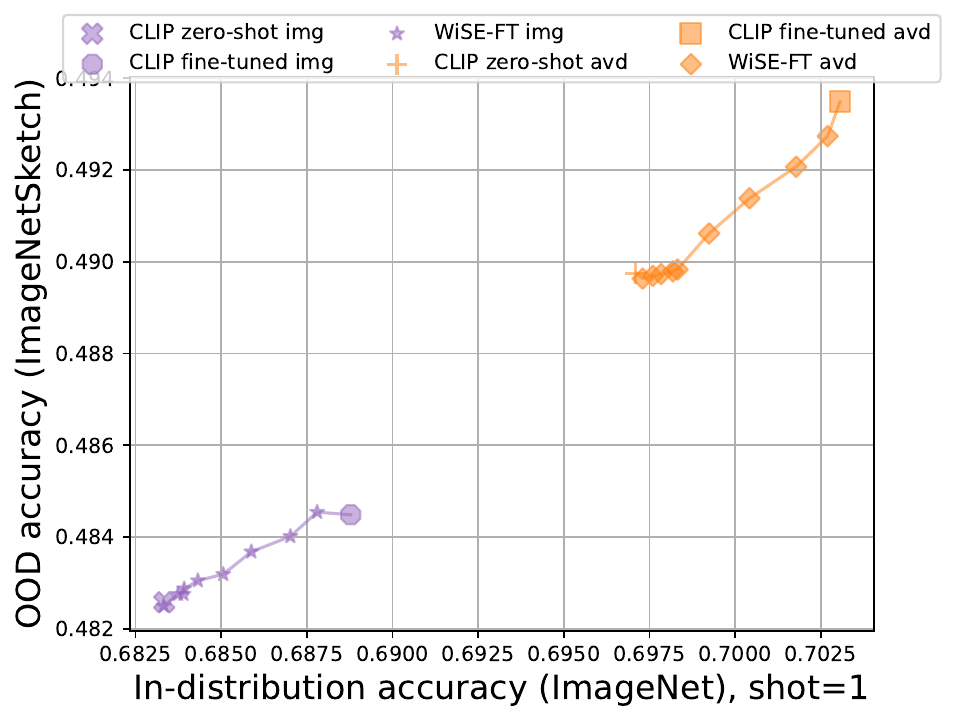}
     \end{subfigure}
     \begin{subfigure}[b]{0.3\textwidth}
         \centering
         \includegraphics[width=\textwidth]{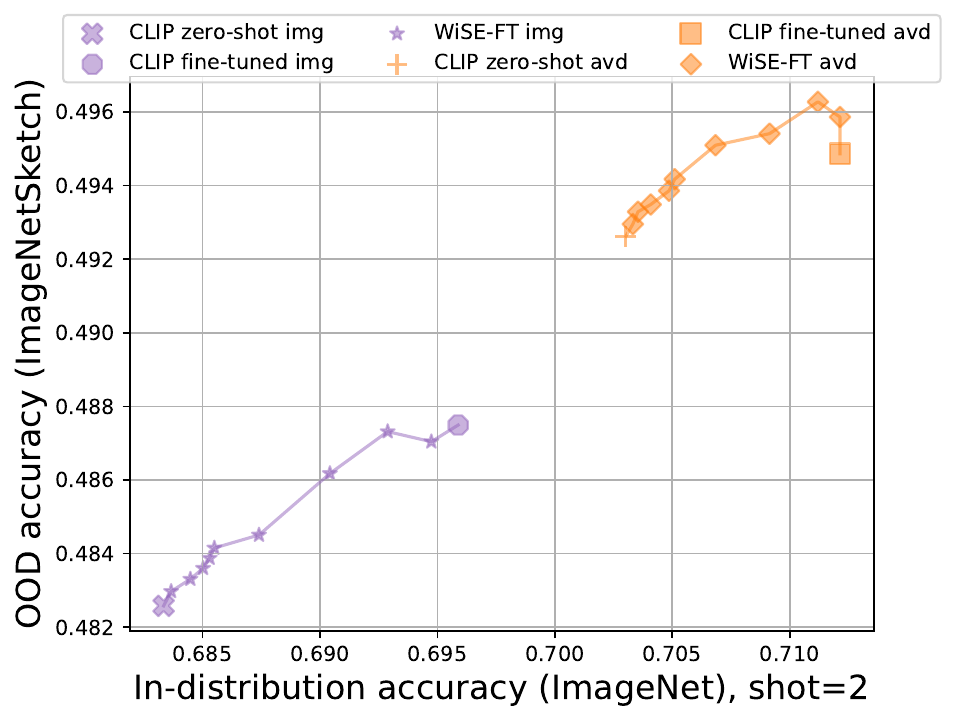}
     \end{subfigure}
     \begin{subfigure}[b]{0.3\textwidth}
         \centering
         \includegraphics[width=\textwidth]{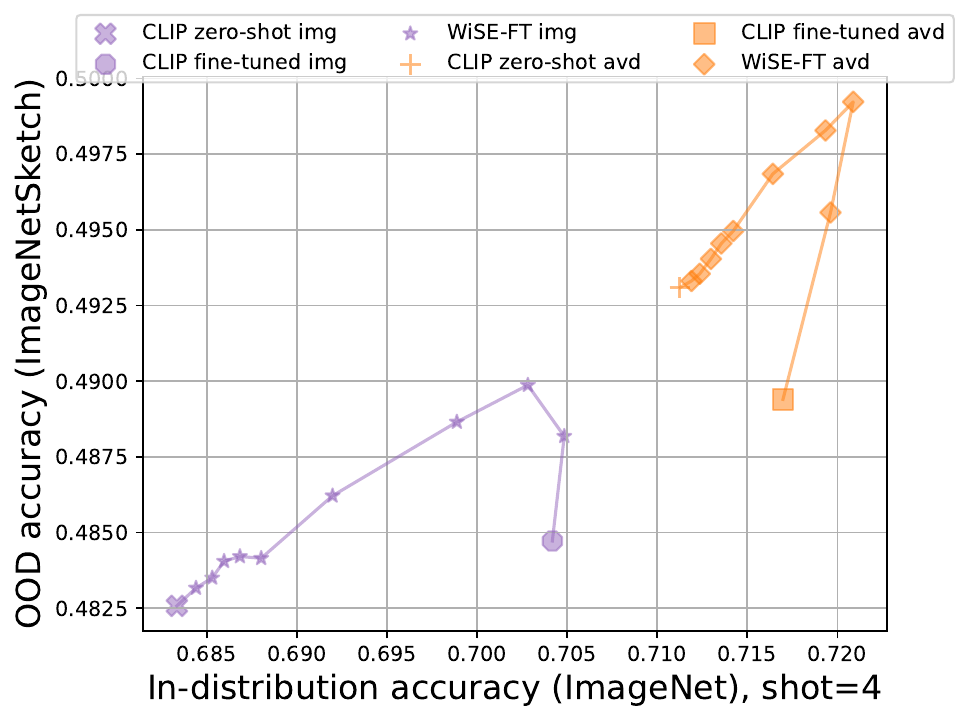}
     \end{subfigure}
    \caption{ID-OOD curves of WISE-FT vs WISE-SLR on IN-Sketch. $k=1,2,4$.}
        \label{fig:id_ood_curve_in_sketch_ft}
\end{figure}

\begin{figure}
     \centering
     \begin{subfigure}[b]{0.3\textwidth}
         \centering
         \includegraphics[width=\textwidth]{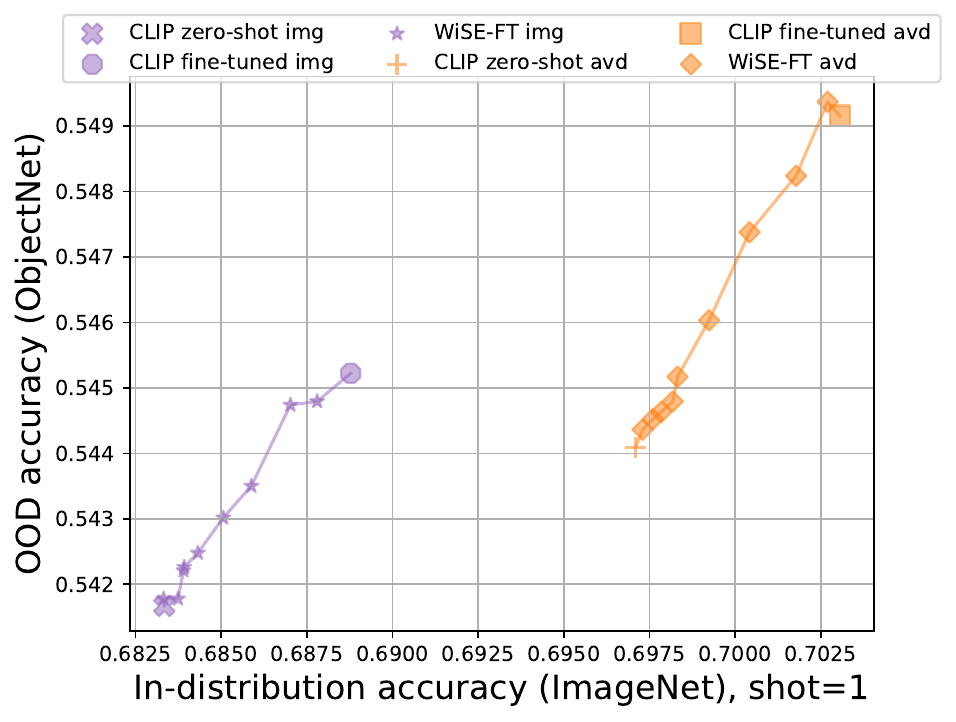}
     \end{subfigure}
     \begin{subfigure}[b]{0.3\textwidth}
         \centering
         \includegraphics[width=\textwidth]{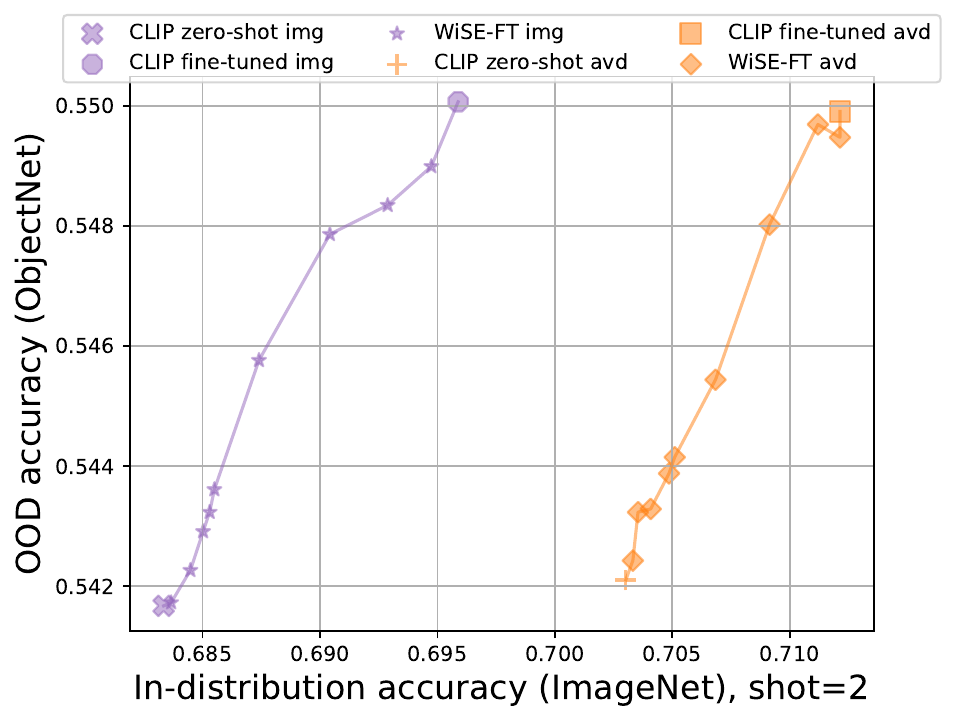}
     \end{subfigure}
     \begin{subfigure}[b]{0.3\textwidth}
         \centering
         \includegraphics[width=\textwidth]{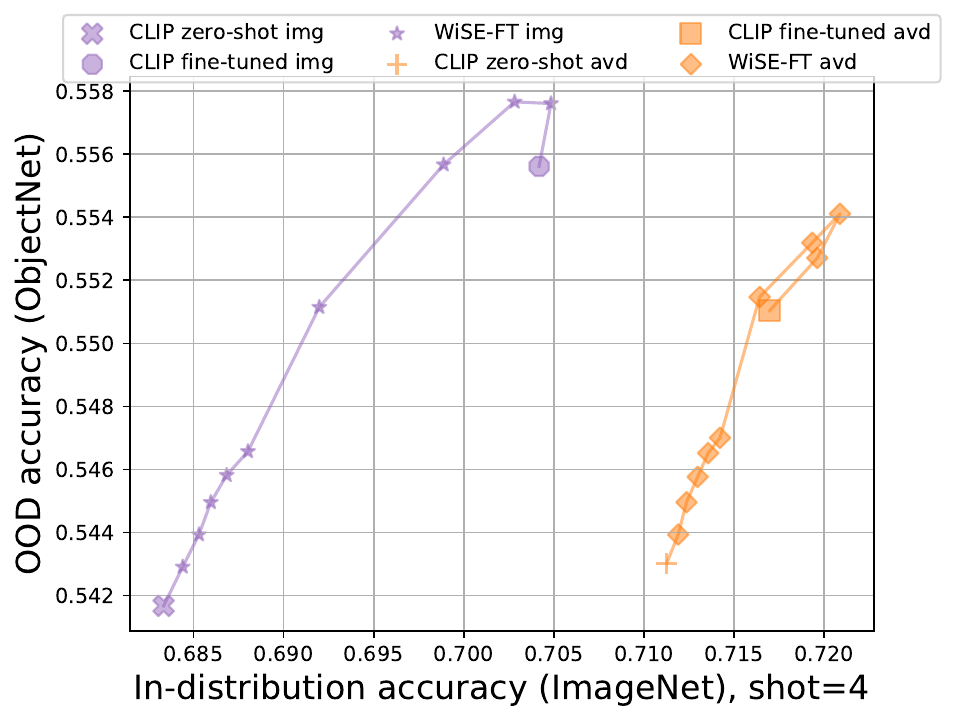}
     \end{subfigure}
    \caption{ID-OOD curves of WISE-FT vs WISE-SLR on ObjectNet. $k=1,2,4$.}
        \label{fig:id_ood_curve_in_object_ft}
\end{figure}

The dataset-wise ID-OOD curves of WISE-FT vs WISE-SLR on IN-A, IN-R, IN-V2, IN-Sketch, and ObjectNet are listed in \cref{fig:id_ood_curve_in_a_ft,fig:id_ood_curve_in_r_ft,fig:id_ood_curve_in_v2_ft,fig:id_ood_curve_in_sketch_ft,fig:id_ood_curve_in_object_ft}, respectively.

%%%%%%%%%%%%%%%%%%%. IMAGENET-A %%%%%%%%%%%%%%%%%%%%%

%%%%%%%%%%%%%%%%%%%. IMAGENET-R %%%%%%%%%%%%%%%%%%%%%

%%%%%%%%%%%%%%%%%%%. IMAGENET-V2 %%%%%%%%%%%%%%%%%%%%%

%%%%%%%%%%%%%%%%%%%. IMAGENET-SKETCH %%%%%%%%%%%%%%%%%%%%%

%%%%%%%%%%%%%%%%%%%. IMAGENET-A %%%%%%%%%%%%%%%%%%%%%

The detailed accuracies of WISE-FT vs WISE-SLR with different choices of $\alpha$ are given in \cref{table:detailed_wiseft_alpha} (ID) and \cref{table:detailed_wiseft_alpha_ood} (OOD). $\alpha=0$ corresponds to zero-shot accuracy, and $\alpha=1$ corresponds to full fine-tuned model. The results in the same setting with only the last linear layer trained are presented in \cref{table:fs_comparison_alpha} (ID) and \cref{table:fs_comparison_alpha_ood} (OOD). 

\begin{table*}[tb]
\centering
\resizebox{\linewidth}{!}{
 \begin{tabular}{c|cc|cc|cc|cc|cc|cc} 
 \midrule
 Shots & \multicolumn{2}{c|}{$k=1$}  &  \multicolumn{2}{c|}{$k=2$} & \multicolumn{2}{c|}{$k=4$}&  \multicolumn{2}{c|}{$k=8$} &  \multicolumn{2}{c|}{$k=16$} &  \multicolumn{2}{c}{$k=32$} \\
 \midrule
 \diagbox{$\alpha$}{Methods} & LP & AVD & LP & AVD  & LP & AVD &  LP & AVD &  LP & AVD &  LP & AVD \\
 \midrule
0.0000 & 68.78 & 69.53 & 68.78 & 69.53 & 68.78 & 69.53 & 68.78 & 69.53 & 68.78 & 69.53 & 68.78 & 69.53 \\
0.0001 & 68.79 & 69.54 & 68.83 & 69.54 & 68.85 & 69.54 & 68.87 & 69.55 & 68.92 & 69.55 & 68.94 & 69.55 \\
0.0002 & 68.81 & 69.54 & 68.88 & 69.55 & 68.91 & 69.55 & 68.93 & 69.56 & 69.02 & 69.56 & 69.07 & 69.56 \\
0.0004 & 68.86 & 69.56 & 68.96 & 69.58 & 69.03 & 69.58 & 69.09 & 69.59 & 69.24 & 69.59 & 69.35 & 69.59 \\
0.0008 & 68.94 & 69.59 & 69.14 & 69.63 & 69.26 & 69.64 & 69.40 & 69.64 & 69.65 & 69.66 & 69.84 & 69.68 \\
0.0016 & 69.06 & 69.62 & 69.38 & 69.72 & 69.63 & 69.74 & 69.93 & 69.79 & 70.36 & 69.79 & 70.70 & 69.83 \\
0.0032 & 69.22 & 69.72 & 69.73 & 69.86 & 70.17 & 69.93 & 70.73 & 70.00 & 71.40 & 70.07 & 72.01 & 70.08 \\
0.0063 & 68.99 & 69.81 & 69.69 & 70.06 & 70.71 & 70.22 & 71.54 & 70.40 & 72.52 & 70.50 & 73.38 & 70.51 \\
0.0126 & 67.31 & 69.83 & 68.04 & 70.33 & 70.35 & 70.75 & 71.65 & 71.09 & 73.10 & 71.25 & 74.22 & 71.27 \\
0.0251 & 62.15 & 69.26 & 63.91 & 70.33 & 68.12 & 71.18 & 70.42 & 71.88 & 72.45 & 72.23 & 74.12 & 72.36 \\
0.0501 & 53.51 & 66.75 & 57.69 & 69.30 & 64.35 & 71.17 & 68.18 & 72.37 & 71.10 & 73.08 & 73.30 & 73.44 \\
0.1000 & 44.41 & 61.19 & 51.99 & 66.38 & 60.59 & 69.99 & 65.81 & 71.96 & 69.62 & 73.45 & 72.46 & 74.26 \\
0.2000 & 37.91 & 53.68 & 47.94 & 62.06 & 57.59 & 67.75 & 64.13 & 70.95 & 68.60 & 73.09 & 71.83 & 74.34 \\
0.3000 & 35.35 & 49.40 & 46.43 & 59.57 & 56.41 & 66.32 & 63.42 & 70.15 & 68.19 & 72.62 & 71.58 & 74.21 \\
0.4000 & 34.05 & 46.60 & 45.63 & 57.93 & 55.83 & 65.43 & 63.06 & 69.52 & 67.99 & 72.22 & 71.44 & 74.06 \\
0.5000 & 33.22 & 44.72 & 45.12 & 56.82 & 55.44 & 64.80 & 62.82 & 69.14 & 67.85 & 72.00 & 71.33 & 73.94 \\
0.6000 & 32.67 & 43.44 & 44.79 & 55.98 & 55.18 & 64.30 & 62.66 & 68.89 & 67.75 & 71.81 & 71.27 & 73.85 \\
0.7000 & 32.27 & 42.47 & 44.52 & 55.39 & 54.99 & 63.92 & 62.54 & 68.68 & 67.70 & 71.67 & 71.23 & 73.77 \\
0.8000 & 31.96 & 41.71 & 44.31 & 54.87 & 54.85 & 63.62 & 62.45 & 68.51 & 67.62 & 71.57 & 71.20 & 73.73 \\
0.9000 & 31.69 & 41.11 & 44.17 & 54.47 & 54.74 & 63.39 & 62.39 & 68.36 & 67.59 & 71.48 & 71.17 & 73.71 \\
1.0000 & 31.51 & 40.56 & 44.06 & 54.16 & 54.66 & 63.19 & 62.33 & 68.23 & 67.55 & 71.40 & 71.15 & 73.67 \\
 \bottomrule
 \end{tabular}
 }
 \caption{Accuracies on ImageNet with difference choices of $\alpha$. We compare LP vs SLR-AVD.}
 \label{table:fs_comparison_alpha}
\end{table*}

\begin{table*}[tb]
\centering
\resizebox{\linewidth}{!}{
 \begin{tabular}{c|cc|cc|cc|cc|cc|cc} 
 \midrule
 Shots & \multicolumn{2}{c|}{$k=1$}  &  \multicolumn{2}{c|}{$k=2$} & \multicolumn{2}{c|}{$k=4$}&  \multicolumn{2}{c|}{$k=8$} &  \multicolumn{2}{c|}{$k=16$} &  \multicolumn{2}{c}{$k=32$} \\
 \midrule
 \diagbox{$\alpha$}{Methods} & LP & AVD & LP & AVD  & LP & AVD &  LP & AVD &  LP & AVD &  LP & AVD \\
 \midrule
0.0000 & 58.66 & 59.03 & 58.66 & 59.03 & 58.66 & 59.03 & 58.66 & 59.03 & 58.66 & 59.03 & 58.66 & 59.03 \\
0.0001 & 58.67 & 59.03 & 58.68 & 59.03 & 58.69 & 59.04 & 58.70 & 59.04 & 58.70 & 59.04 & 58.71 & 59.04 \\
0.0002 & 58.68 & 59.02 & 58.68 & 59.04 & 58.70 & 59.04 & 58.71 & 59.04 & 58.70 & 59.04 & 58.73 & 59.04 \\
0.0004 & 58.69 & 59.03 & 58.70 & 59.04 & 58.71 & 59.04 & 58.73 & 59.03 & 58.75 & 59.05 & 58.81 & 59.05 \\
0.0008 & 58.70 & 59.03 & 58.75 & 59.06 & 58.81 & 59.06 & 58.84 & 59.06 & 58.88 & 59.06 & 58.99 & 59.06 \\
0.0016 & 58.72 & 59.06 & 58.81 & 59.08 & 58.90 & 59.09 & 59.03 & 59.11 & 59.07 & 59.08 & 59.23 & 59.11 \\
0.0032 & 58.69 & 59.07 & 58.77 & 59.13 & 58.97 & 59.18 & 59.15 & 59.19 & 59.20 & 59.16 & 59.47 & 59.21 \\
0.0063 & 58.30 & 59.10 & 58.27 & 59.14 & 58.79 & 59.23 & 58.89 & 59.31 & 58.98 & 59.28 & 59.40 & 59.37 \\
0.0126 & 56.77 & 58.90 & 56.41 & 59.16 & 57.40 & 59.30 & 57.56 & 59.45 & 57.79 & 59.49 & 58.34 & 59.59 \\
0.0251 & 52.55 & 58.09 & 51.93 & 58.70 & 54.30 & 59.14 & 54.81 & 59.38 & 55.47 & 59.55 & 56.28 & 59.78 \\
0.0501 & 45.05 & 55.46 & 45.65 & 57.11 & 49.68 & 58.20 & 51.10 & 58.73 & 52.54 & 59.13 & 53.86 & 59.76 \\
0.1000 & 36.63 & 50.25 & 39.71 & 53.88 & 45.30 & 56.12 & 47.77 & 57.21 & 50.03 & 58.05 & 51.87 & 59.25 \\
0.2000 & 30.30 & 43.39 & 35.56 & 49.45 & 42.08 & 53.33 & 45.47 & 55.21 & 48.38 & 56.46 & 50.61 & 58.29 \\
0.3000 & 27.83 & 39.70 & 33.94 & 47.00 & 40.79 & 51.76 & 44.55 & 54.02 & 47.73 & 55.60 & 50.12 & 57.70 \\
0.4000 & 26.59 & 37.37 & 33.09 & 45.44 & 40.10 & 50.76 & 44.09 & 53.25 & 47.39 & 55.02 & 49.88 & 57.31 \\
0.5000 & 25.82 & 35.84 & 32.57 & 44.43 & 39.67 & 50.12 & 43.79 & 52.74 & 47.18 & 54.63 & 49.71 & 57.04 \\
0.6000 & 25.28 & 34.73 & 32.22 & 43.70 & 39.40 & 49.63 & 43.59 & 52.36 & 47.03 & 54.37 & 49.62 & 56.82 \\
0.7000 & 24.87 & 33.94 & 31.99 & 43.16 & 39.18 & 49.25 & 43.45 & 52.07 & 46.92 & 54.16 & 49.55 & 56.67 \\
0.8000 & 24.59 & 33.33 & 31.80 & 42.72 & 39.03 & 48.95 & 43.34 & 51.86 & 46.83 & 54.02 & 49.49 & 56.54 \\
0.9000 & 24.36 & 32.85 & 31.65 & 42.38 & 38.91 & 48.71 & 43.26 & 51.67 & 46.77 & 53.88 & 49.44 & 56.45 \\
1.0000 & 24.17 & 32.42 & 31.53 & 42.08 & 38.80 & 48.51 & 43.19 & 51.54 & 46.72 & 53.79 & 49.41 & 56.35 \\
 \bottomrule
 \end{tabular}
 }
 \caption{Accuracies on ImageNet variations with difference choices of $\alpha$. We compare LP vs SLR-AVD. The results are averaged over all 5 ImageNet variations.}
 \label{table:fs_comparison_alpha_ood}
\end{table*}

\begin{table}[tb]
\centering
\resizebox{\columnwidth}{!}{
 \begin{tabular}{c|c|c|c|c|c|c} 
 \toprule
 Shot&  \multicolumn{2}{c|}{$k=1$} & \multicolumn{2}{c|}{$k=2$} & \multicolumn{2}{c}{$k=4$}\\
 \midrule
\diagbox{$\alpha$}{Method}& WISE-FT & WISE-SLR & WISE-FT & WISE-SLR &WISE-FT & WISE-SLR\\
\midrule
0.00 & 58.39 & 59.07    & 58.39 & 59.21    & 58.39 & 59.33 \\
0.02 & 58.40 & 59.09    & 58.40 & 59.22    & 58.45 & 59.39 \\
0.04 & 58.40 & 59.11    & 58.44 & 59.27    & 58.53 & 59.45 \\
0.06 & 58.42 & 59.11    & 58.46 & 59.31    & 58.62 & 59.49 \\
0.08 & 58.42 & 59.12    & 58.48 & 59.33    & 58.69 & 59.58 \\
0.10 & 58.44 & 59.14    & 58.51 & 59.38    & 58.73 & 59.63 \\
0.20 & 58.46 & 59.20    & 58.65 & 59.48    & 59.07 & 59.97 \\
0.40 & 58.50 & 59.29    & 58.82 & 59.68    & 59.40 & 60.24 \\
0.60 & 58.55 & 59.38    & 58.97 & 59.80    & 59.60 & 60.37 \\
0.80 & 58.56 & 59.44    & 58.98 & 59.75    & 59.68 & 60.19 \\
1.00 & 58.58 & 59.49    & 58.98 & 59.74    & 59.50 & 59.88 \\
\bottomrule
 \end{tabular}
 }
  \caption{Accuracies on ImageNet variations with difference choice of $\alpha$. We compare WISE-FT to WISE-SLR. The results are averaged over 5 ImageNet variations.}
 \label{table:detailed_wiseft_alpha_ood}
\end{table}

\begin{table}[tb]
\centering
\resizebox{\columnwidth}{!}{
 \begin{tabular}{c|c|c|c|c|c|c} 
 \toprule
 Shot&  \multicolumn{2}{c|}{$k=1$} & \multicolumn{2}{c|}{$k=2$} & \multicolumn{2}{c}{$k=4$}\\
 \midrule
\diagbox{$\alpha$}{Method}& WISE-FT & WISE-SLR & WISE-FT & WISE-SLR &WISE-FT & WISE-SLR\\
\midrule
0.00 & 68.33 & 69.71    & 68.33 & 70.30    & 68.33 & 71.13 \\
0.02 & 68.33 & 69.73    & 68.37 & 70.33    & 68.44 & 71.19 \\
0.04 & 68.37 & 69.76    & 68.45 & 70.35    & 68.53 & 71.24 \\
0.06 & 68.39 & 69.78    & 68.50 & 70.41    & 68.60 & 71.30 \\
0.08 & 68.39 & 69.82    & 68.53 & 70.49    & 68.68 & 71.36 \\
0.10 & 68.43 & 69.83    & 68.55 & 70.51    & 68.80 & 71.42 \\
0.20 & 68.51 & 69.92    & 68.74 & 70.68    & 69.20 & 71.64 \\
0.40 & 68.59 & 70.04    & 69.04 & 70.91    & 69.89 & 71.93 \\
0.60 & 68.70 & 70.18    & 69.29 & 71.12    & 70.28 & 72.09 \\
0.80 & 68.78 & 70.27    & 69.47 & 71.21    & 70.48 & 71.96 \\
1.00 & 68.88 & 70.31    & 69.59 & 71.21    & 70.42 & 71.70 \\
\bottomrule
 \end{tabular}
 }
 \caption{Accuracies on ImageNet with difference choice of $\alpha$. We compare WISE-FT to WISE-SLR.}
 \label{table:detailed_wiseft_alpha}
\end{table}

\paragraph{Choosing $\gamma$ and LLM prompting} We consider another prompt ``Give me 100 useful visual features for distinguishing \{\} in a photo'', and use it with frequency penalty (FP) $0$ in $\textcircled{1}$, FP $0.1$ in $\textcircled{2}$. $\textcircled{3}$ uses the GPT3 prompts in the main text with $0$ FP. Unless otherwise specified, other experiments use $\gamma=\frac{1}{M_c+1}$ and FP $0.1$, and the GPT prompts in the main text. We find that the GPT3 prompt itself does not matter as much as FP -- it is more important to generate a more diverse set of VD. Note in the main text we set $\gamma=5$, this is because on ImageNet it is hard to guarantee the same $M_c$ across classes (due to an excess number of classes), hence we use a large $\gamma$ to enforce ZS-AVD relies mostly on the strong class prompts. In this ablation study, we enforce GPT to give $100$ VDs per class so we can simply average over them.

\begin{table}[tb]
\small
\centering
 \caption{ZS ablation on $\gamma$ and GPT prompts.}
 % \resizebox{\columnwidth}{!}{
 \begin{tabular}{c|c|c|c|c|c|c} 
 \midrule
 $\gamma$ or prompts & $1/(M_c+1)$  &  1 & 5  &\textcircled{1} &\textcircled{2} &\textcircled{3} \\
 \midrule
CIFAR10 &  91.51 &  91.19  & 91.16 &91.25 &  91.42  & 90.44\\
CIFAR10.1 & 86.35 &  85.90 & 85.90  & 85.90 &  85.60 & 85.40 \\
CIFAR10.2 & 83.80 &  83.10 & 83.10   & 83.20 &  84.20 & 82.50  \\
 \bottomrule
 \end{tabular}
 \label{table:ablationllm}
 % }
\end{table}

\end{document}